\documentclass[a4paper,fleqn]{cas-dc}
\usepackage[numbers,sort&compress]{natbib}
\usepackage{microtype}

\hypersetup{pdftitle={PolMamba: Long-wave Infrared Polarimetric Remote Sensing Image Fusion for Vehicle Detection},pdfauthor={}}
\ExplSyntaxOn
\cs_set:Npn \__first_footerline: {\group_begin:\small\rmfamily\itshape Preprint~prepared~using~the~Elsevier~CAS~template\group_end:}
\RenewDocumentEnvironment{Abstract}{o}
{
  \group_begin:
  \IfNoValueTF{#1}{}{\tex_gdef:D\abstractname{#1}}
  \parindent=0pt
  \box_if_empty:NTF\g_stm_key_box
    {\leftskip=.35\textwidth}
    {
      \dim_gset:Nn\l_tmpa_dim{\box_ht:N\g_stm_key_box}
      \dim_gadd:Nn\l_tmpa_dim{\box_dp:N\g_stm_key_box}
      \leftskip=.35\textwidth
      \hspace*{-.35\textwidth}
      \noindent\hbox_to_wd:nn{0pt}{\box\g_stm_key_box\hss}
      \skip_vertical:n{-\l_tmpa_dim}
    }
  \noindent\abstractname\par
  \skip_vertical:n{-4pt}
  \noindent\rule{.65\textwidth}{.2pt}\par\footnotesize
  \ignorespaces\everypar{\parindent=1.5em}
}
{\par\group_end:}
\ExplSyntaxOff

\ExplSyntaxOn
\cs_new_eq:NN \cas_original_fig_caption:nn \__make_fig_caption:nn
\cs_set:Npn \__make_fig_caption:nn #1#2
{
	\group_begin:
	\cs_set_protected:Npn \sffamily
	{ \fontfamily{ptm}\selectfont }
	\cas_original_fig_caption:nn {#1}{#2}
	\group_end:
}
\ExplSyntaxOff

\ExplSyntaxOn
\cs_new_eq:NN \cas_original_tbl_caption:nn \__make_tbl_caption:nn
\cs_set:Npn \__make_tbl_caption:nn #1#2
{
	\group_begin:
	\cs_set_protected:Npn \sffamily
	{ \fontfamily{ptm}\selectfont }
	\cas_original_tbl_caption:nn {#1}{#2}
	\group_end:
}
\ExplSyntaxOff

\AddToHook{env/tabular/begin}{\fontfamily{ptm}\selectfont}
\AddToHook{env/tabular*/begin}{\fontfamily{ptm}\selectfont}

\begin{document}
\let\WriteBookmarks\relax
\shorttitle{LWIR polarimetric remote sensing image fusion for vehicle detection}
\shortauthors{}
\title[mode=title]{PolMamba: Long-wave Infrared Polarimetric Remote Sensing Image Fusion for Vehicle Detection}
\author[1]{Wenjuan Zhang}
\author[2,3]{Xiangrui Peng}
\author[2]{Zhen Li}
\cormark[1]
\ead{lizhen02@aircas.ac.cn}
\author[2,3]{Caimeng Yi}
\author[2,3]{Xuesong Ma}
\author[2]{Jianwei Wang}
\author[1]{Jie Pan}
\affiliation[1]{organization={State Key Laboratory of Remote Sensing and Digital Earth, Aerospace Information Research Institute, Chinese Academy of Sciences},city={Beijing},country={China}}
\affiliation[2]{organization={Aerospace Information Research Institute, Chinese Academy of Sciences},city={Beijing},country={China}}
\affiliation[3]{organization={University of Chinese Academy of Sciences},city={Beijing},country={China}}
\cortext[1]{Corresponding author}
\begin{abstract}
Blurred vehicle contours and low contrast limit detection in infrared remote sensing imagery, while fusion can introduce polarization noise and incompletely integrate complementary information. We propose PolMamba, a Mamba-based infrared polarimetric remote sensing image fusion model. Infrared polarization provides information about surface materials and geometry that complements intensity imagery. PolMamba combines long-wave infrared total intensity ($S_0$) and the degree of linear polarization (DoLP), using a physics-inspired polarization-response proxy and spatial structural continuity to generate guidance features. Bidirectional state space scanning along horizontal, vertical, and both diagonal orientations enables image-wide cross-modal interaction. Prior-guided multi-window structural similarity, reconstruction constraints, and a relative lower-bound loss for polarization saliency jointly preserve background structures and target polarization features while suppressing noise. We also construct AIR-LWIR-Pol, containing 1,901 accurately registered image pairs acquired by an airborne division-of-aperture long-wave infrared polarimetric system, and evaluate downstream vehicle detection using YOLO11. PolMamba ranks first on four fusion metrics: entropy, standard deviation, structural similarity, and NIMA. Vehicle detection precision and $\mathrm{mAP}_{50}$ reach 60.7\% and 57.4\%, improving on infrared intensity alone by 5.6 and 1.5 percentage points, respectively. These results demonstrate the value of combining physical priors with state space modeling for long-wave infrared polarization fusion and vehicle detection in complex remote sensing scenes.
\end{abstract}
\begin{keywords}[Dataset link]
    {\urlstyle{same}\url{https://github.com/peng-xiangrui/AIR-LWIR-Pol-dataset}}
    \par\medskip
    \textit{Model link:}\par
    {\urlstyle{same}\url{https://github.com/peng-xiangrui/PolMamba}}
    \par\medskip
    \textit{Keywords:}\par
    Long-wave infrared polarimetry \sep Image fusion \sep Physical priors \sep State space models \sep Vehicle detection
\end{keywords}

\maketitle
\section{Introduction}\label{sec:introduction}
Thermal infrared remote sensing directly captures thermal radiation emitted by surface objects and is widely used in Earth observation, urban monitoring, disaster assessment, and target perception. Conventional infrared intensity imaging relies primarily on radiance differences between targets and their surroundings. Small temperature differences, atmospheric disturbances, and complex backgrounds can reduce target--background contrast and make detection and recognition more difficult.

Infrared polarimetric imaging provides complementary information about surface materials, roughness, and geometry. Infrared intensity $S_0$ characterizes the distribution of total infrared radiation, whereas the degree of linear polarization (DoLP) describes the proportion of the linearly polarized component in the total intensity. These complementary imaging mechanisms can improve target--background separability in scenes with low thermal contrast or complex backgrounds.

Long-wave infrared polarization responses are often weak, and DoLP images can contain unstable structures, including isolated high responses and random background fluctuations. A high DoLP response may arise from polarization noise rather than a reliable target, while weak responses can contain spatially continuous target contours. Fusion therefore requires both polarization feature extraction and spatial prior estimation to distinguish structured polarization information from isolated responses.

Deep models, including convolutional neural networks (CNNs), generative adversarial networks (GANs), autoencoders, and Transformers, have been applied to infrared and polarization image fusion. Learned feature representations improve the extraction of complementary information, while polarization priors and noise-aware mechanisms help preserve polarization details and suppress noise. However, the different physical meanings of infrared intensity and DoLP make it difficult for a purely data-driven mapping to determine whether a polarization response has stable spatial support. Common objectives, such as $L_1$, mean squared error (MSE), and structural similarity (SSIM), constrain pixel or structural consistency without explicitly modeling polarization-response priors.

High-resolution remote sensing also requires effective feature interaction over long distances. CNNs extract local textures, but their spatial interaction range is constrained by their receptive fields. Transformers establish image-wide relationships through self-attention, although not all distant features are equally relevant to fusion. The first law of geography suggests that nearby objects tend to be more closely related, motivating spatially organized feature interaction. Standard self-attention also incurs approximately quadratic computational and memory costs with respect to the number of spatial tokens. State space models (SSMs) provide a sequence-scanning mechanism for propagating features along spatially ordered paths, with approximately linear complexity in sequence length. This motivates adapting state space scans to the horizontal, vertical, and oblique contours found in remote sensing imagery.

\section{Motivation}\label{sec:motivation}
We develop a unified fusion framework that connects polarization priors, long-range cross-modal interaction, and physics-inspired optimization. The main contributions are as follows:
\begin{enumerate}
\item We introduce a physics-guided polarization prior enhancement module (PPREM) that incorporates polarization imaging relationships into feature enhancement and network optimization. By jointly modeling polarization information and spatial structural continuity, PPREM produces continuous guidance to suppress isolated responses while preserving weak polarization information with spatial support.
\item We develop a multidirectional Cross-Mamba fusion mechanism that propagates features along horizontal, vertical, and both diagonal orientations. Prior guidance regulates the interaction between infrared intensity and polarization features to strengthen cross-modal spatial representation.
\item We formulate a one-sided lower-bound loss for infrared polarization image fusion. It penalizes a fused response only when it falls below a polarization saliency reference, coordinating the integration of intensity and DoLP information.
\item We construct AIR-LWIR-Pol, an airborne long-wave infrared polarimetric remote sensing dataset containing intensity and polarization images from diverse scenes, processed image pairs, and target annotations, to support fusion and downstream perception research.
\end{enumerate}

\section{Related Work}\label{sec:related}
\subsection{Multimodal Remote Sensing Image Fusion}
Multimodal image fusion integrates complementary information from different imaging mechanisms to improve scene representation. Early methods rely on manually designed transforms and fusion rules, including region-level fusion\cite{ref5}, multiscale decomposition\cite{ref6}, transform-domain methods\cite{ref7,ref8}, sparse representation\cite{ref9}, and saliency modeling\cite{ref10}. These methods offer explicit mathematical interpretations, but their representation capacity is limited. Complex land-cover distributions and substantial variations in spatial scale further increase the demand for expressive fusion models in remote sensing.

CNNs, Transformers, and hybrid architectures have become important approaches to remote sensing image fusion. CNN-based models extract local textures and edges. He et al.\cite{ref11} improve detail preservation through spatial--spectral modeling, while Ji et al.\cite{ref12} use a pretrained masked autoencoder to enhance feature representation. Transformer-based regression networks and multibranch architectures\cite{ref13,ref14,ref15} extend long-range dependency modeling through self-attention for high-resolution remote sensing fusion. The computational cost of standard self-attention at increasing spatial resolutions motivates more efficient global feature interaction.

Recent Mamba architectures propagate long-range information through state updates with approximately linear complexity in sequence length. Peng et al.\cite{ref16} use a dual-branch U-shaped state space network for cross-modal fusion, and Zhu et al.\cite{ref17} combine Mamba and CNNs to capture global interactions and local details. Visual SSMs commonly unfold images along horizontal and vertical directions, whereas roads, shorelines, and target boundaries occur at diverse orientations. This study examines scan direction design as a means of aligning cross-modal feature propagation with spatial structure.

\subsection{Infrared and Polarization Image Fusion}
Infrared polarization image fusion combines intensity and polarization information to enhance target structures and scene representation. Existing approaches include multiscale modeling\cite{ref18}, low-rank sparse representation\cite{ref19,ref20,ref21}, and Transformer-based cross-modal interaction\cite{ref23}. Polarization-prior-guided and noise-aware methods\cite{ref20,ref22} demonstrate the value of explicitly modeling polarization responses and noise. The next step is to connect this guidance across feature enhancement, cross-modal propagation, and optimization.

Physically, $S_0$ represents total radiative intensity, whereas DoLP is the proportion of linearly polarized radiation. Its response depends on material properties, surface geometry, and observation conditions. In remote sensing images, spatially continuous target contours coexist with isolated high responses and random background fluctuations. Prior guidance and filtering can suppress polarization noise\cite{ref20,ref22}, but excessive smoothing can weaken small targets and fine edges. Effective polarization fusion must preserve these weak structures while reducing noise.

The linearly polarized intensity (PI) derived from the Stokes parameters obeys physical realizability relationships. Common MSE, $L_1$, SSIM, and gradient losses\cite{ref20,ref23} constrain fused images at the pixel or structural level. To retain informative polarization responses during cross-modal interaction and optimization, we use the Stokes relationship between PI, $S_0$, and DoLP to construct a physics-inspired polarization saliency proxy and impose a one-sided relative lower-bound constraint.

\section{Dataset}\label{sec:dataset}
\subsection{Data Acquisition}
\begin{figure}[pos=tp]
\centering
\includegraphics[width=\linewidth,height=.72\textheight,keepaspectratio]{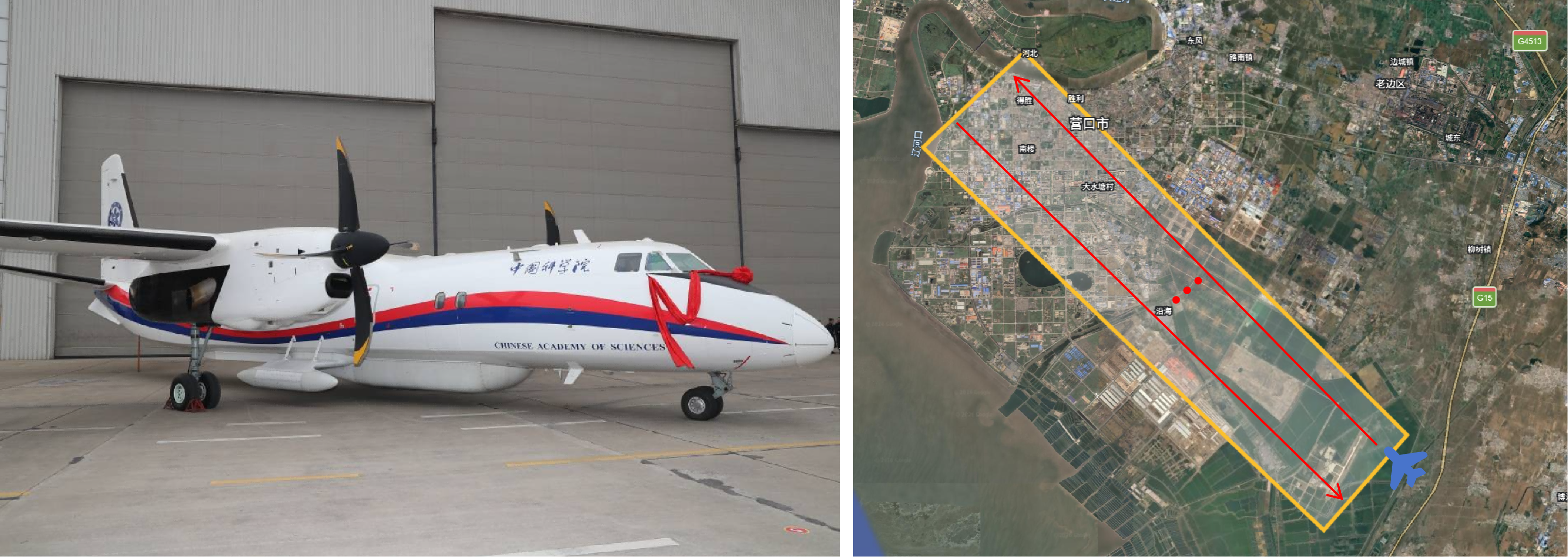}
\caption{Flight lines and the remote sensing aircraft.}
\label{fig:1}
\end{figure}

\begin{figure}[pos=tp]
\centering
\includegraphics[width=\linewidth,height=.72\textheight,keepaspectratio]{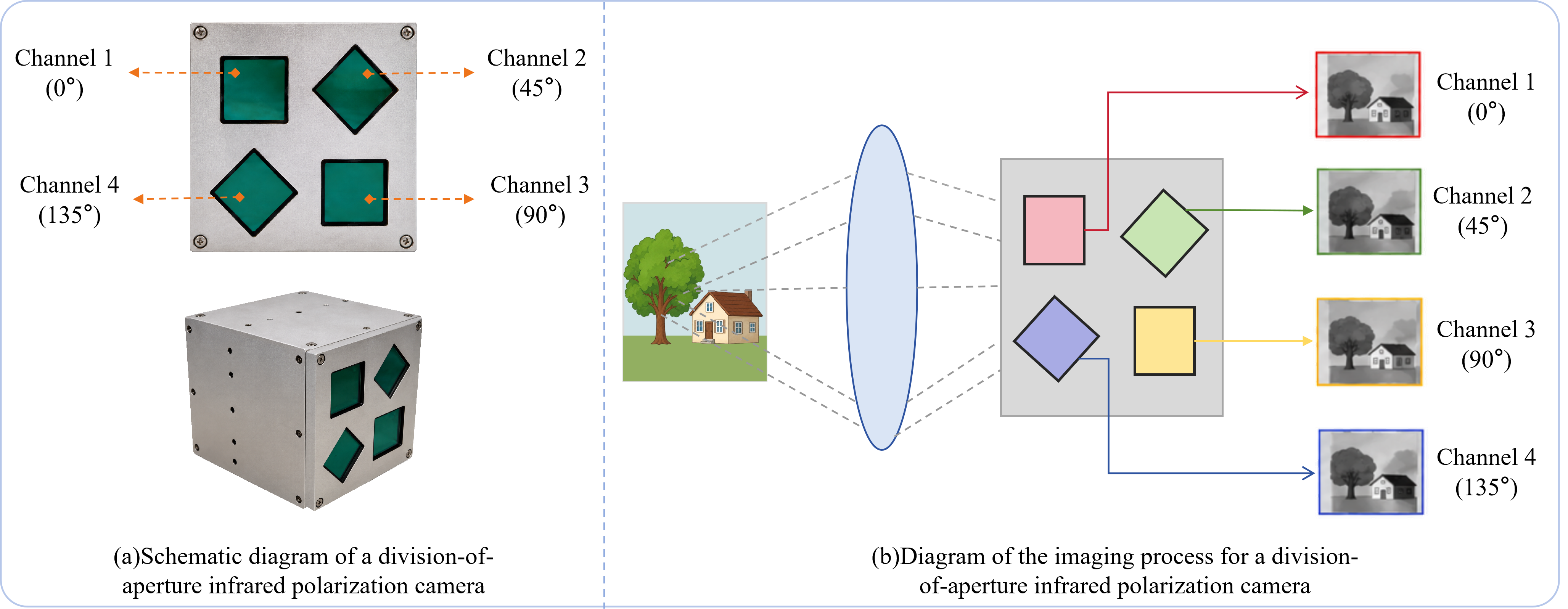}
\caption{Division-of-aperture long-wave infrared polarimetric camera and imaging principle.}
\label{fig:2}
\end{figure}

The experiments use AIR-LWIR-Pol, a long-wave infrared polarimetric remote sensing dataset acquired by a division-of-aperture imaging system aboard the MA60 remote sensing aircraft operated by the Airborne Remote Sensing Center, Chinese Academy of Sciences. Data were collected over Yingkou, Liaoning Province, China, from 14:00 to 19:00 on December 18, 2025, at an altitude of approximately 3,000~m. The survey comprised 16 flight lines with a reported total length of approximately 372.61~m and covered approximately 119.43~km$^2$, including vegetation, bare soil, water bodies, urban areas, and farmland (Fig.~\ref{fig:1}).
The camera operates in the 8--12~$\mu$m spectral range, with a focal length of 55~mm and a detector pixel pitch of 12~$\mu$m, providing a spatial resolution of approximately 0.68~m. As illustrated in Fig.~\ref{fig:2}, the division-of-aperture system simultaneously acquires infrared radiation at polarization orientations of $0^\circ$, $45^\circ$, $90^\circ$, and $135^\circ$.

The raw airborne infrared polarimetric imagery was processed through image registration, relative nonuniformity correction, and annotation to construct AIR-LWIR-Pol.

\subsection{Dataset Description}
\begin{figure*}[pos=tp]
\centering
\includegraphics[width=\linewidth,height=.72\textheight,keepaspectratio]{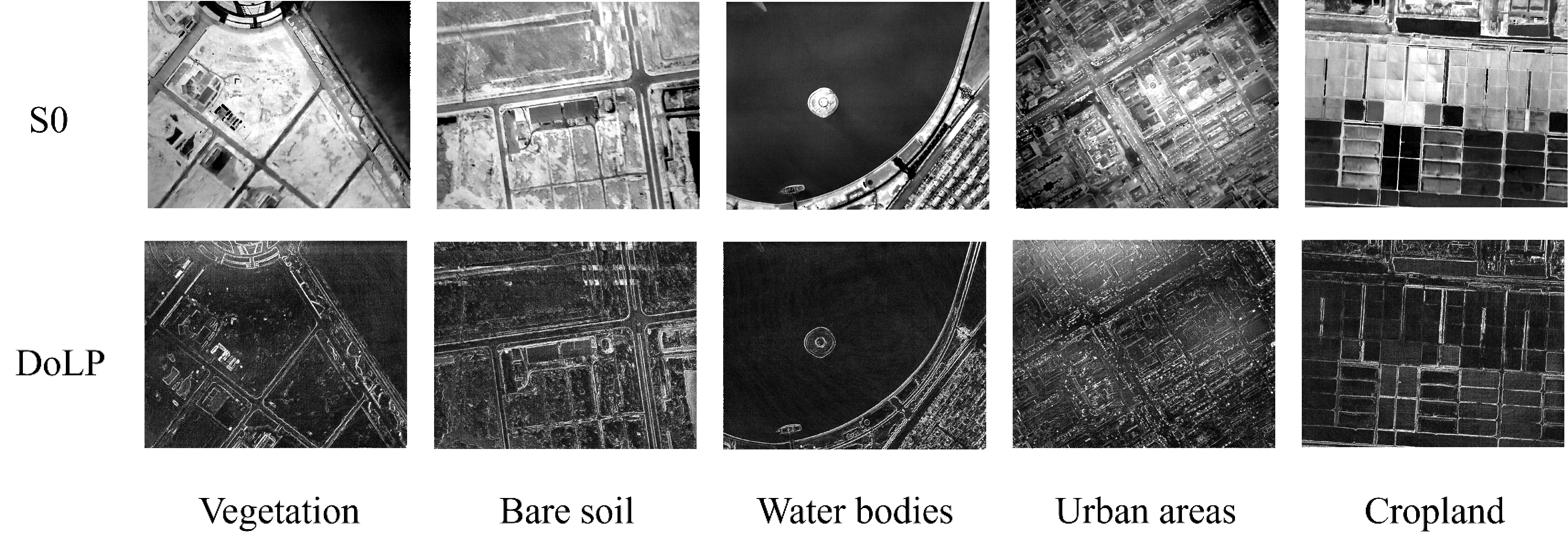}
\caption{Representative land-cover scenes in the AIR-LWIR-Pol dataset.}
\label{fig:3}
\end{figure*}

The original images have a bit depth of 16 bits. Corresponding 8-bit images are generated for network input and visual evaluation using linear contrast stretching with 2\% clipping at each end of the intensity distribution. AIR-LWIR-Pol contains 1,901 registered $S_0$--DoLP pairs of $1280\times1024$ pixels, with a spatial resolution of approximately 0.68~m. As shown in Fig.~\ref{fig:3},the dataset covers five representative land-cover types---vegetation, bare soil, water bodies, urban areas, and farmland---and provides oriented bounding-box annotations for downstream vehicle detection.

\section{Methods}\label{sec:methods}
\subsection{Overall Architecture}
\begin{figure*}[pos=tp]
\centering
\includegraphics[width=\linewidth,height=.72\textheight,keepaspectratio]{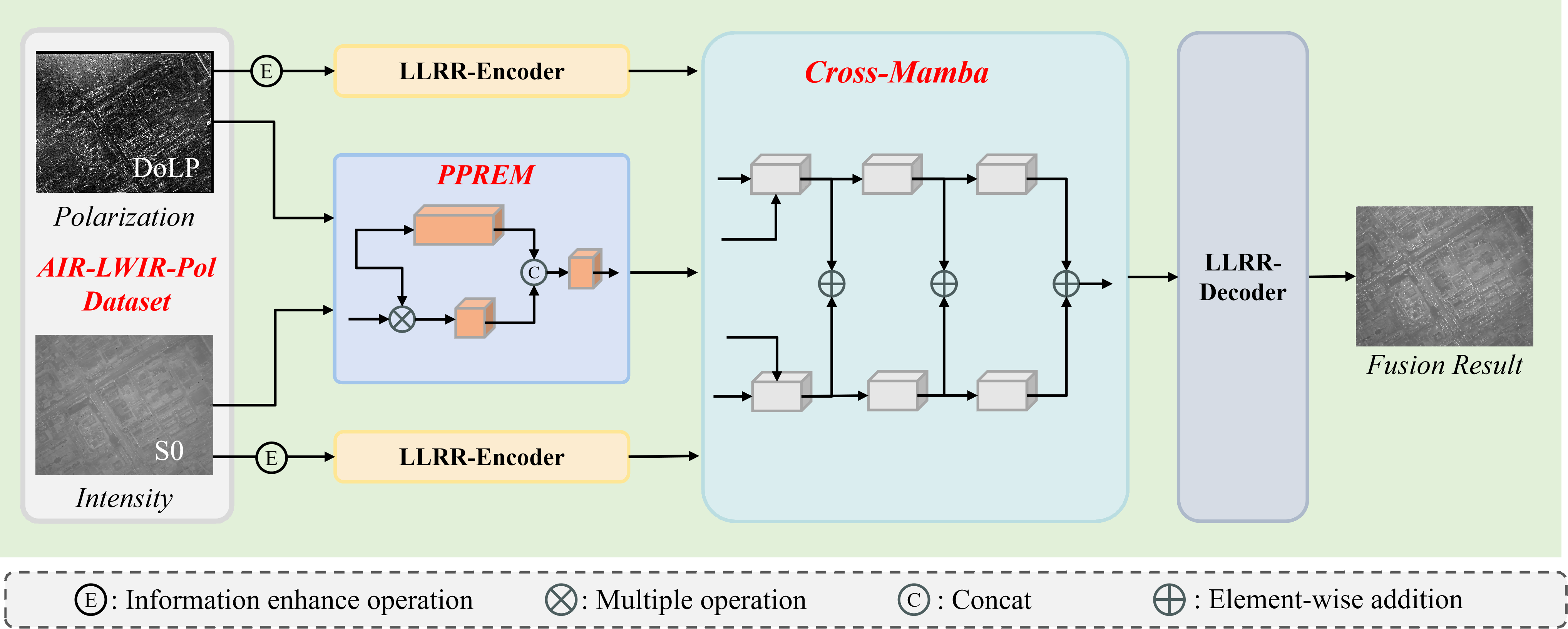}
\caption{Architecture of the proposed infrared polarimetric remote sensing image fusion network.}
\label{fig:4}
\end{figure*}

As shown in Fig.~\ref{fig:4}, PolMamba comprises a dual-branch LLRR encoder, PPREM, Cross-Mamba, and an LLRR decoder. The $S_0$ and DoLP images are independently encoded into shallow modality features. PPREM uses both inputs to construct spatial prior guidance, which regulates long-range cross-modal interaction in Cross-Mamba. The resulting fused features are reconstructed by the decoder to produce the fused image.

\subsection{LLRR-Based Dual-Branch Encoder--Decoder}
\begin{figure}[pos=tp]
\centering
\includegraphics[width=\linewidth,height=.72\textheight,keepaspectratio]{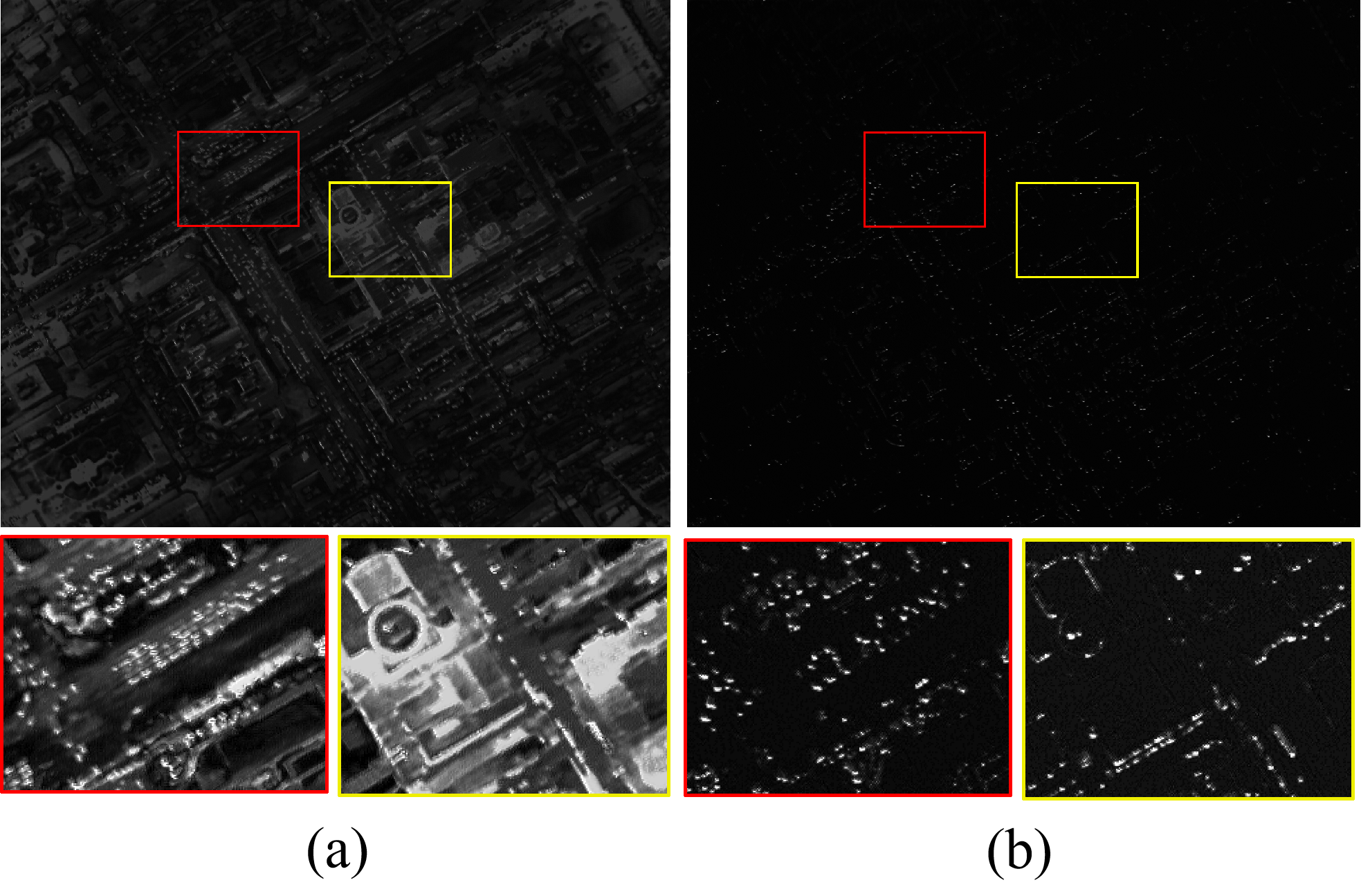}
\caption{Low-rank and sparse characteristics of the source images: (a) low-rank structures in $S_0$; (b) sparse responses in DoLP.}
\label{fig:5}
\end{figure}

Infrared intensity images preserve background textures, such as buildings and ground surfaces, but small vehicles can appear indistinct. DoLP images reveal polarization responses from artificial metallic targets, including vehicles, while their backgrounds are less distinct because background polarization is generally weak. In Fig.~\ref{fig:5}, the yellow-box building region in $S_0$ contains clearer texture than the corresponding DoLP region, motivating a low-rank representation. Conversely, the red-box DoLP region contains prominent vehicle responses, motivating a sparse representation.

We employ two encoders with identical architectures and independent parameters to represent the modalities while limiting interference during shallow feature extraction. Large background regions exhibit strong spatial correlation, whereas target edges, local textures, and polarization structures often form sparse local responses. We therefore adopt the LLRR encoder of Li et al.\cite{ref20} as the shallow feature extractor.
\begin{figure}[pos=tp]
\centering
\includegraphics[width=\linewidth,height=.72\textheight,keepaspectratio]{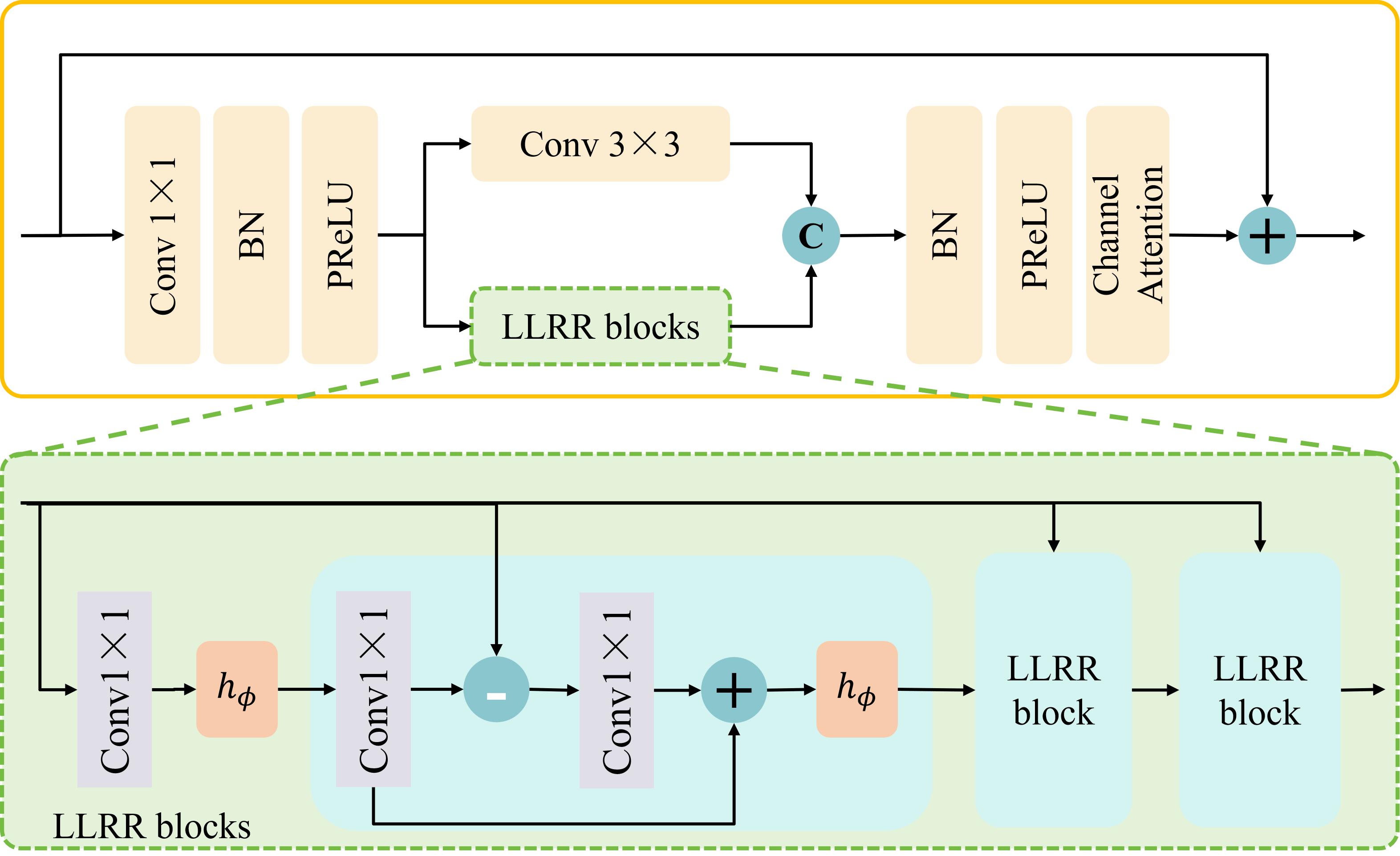}
\caption{Architecture of the LLRR-based encoder--decoder.}
\label{fig:6}
\end{figure}
As shown in Fig.~\ref{fig:6}, the intensity branch produces $F_{S_0}$, primarily representing scene intensity and background structures, while the polarization branch produces $F_{\mathrm{DoLP}}$, emphasizing polarization contours and textures. These features are subsequently passed to Cross-Mamba. During training, images are randomly cropped into $256\times256$ patches to control memory consumption and increase the diversity of local spatial configurations. The LLRR architecture and parameter settings follow Li et al.\cite{ref20}.

\subsection{Physics-Guided Polarization Prior Enhancement Module}
In long-wave infrared remote sensing, DoLP provides information about surface materials and geometry, but informative responses are often weak and accompanied by random fluctuations or isolated high responses. Treating all DoLP responses on an equal footing with $S_0$ can introduce polarization noise or attenuate informative polarization structures when the intensity background dominates.

\begin{figure}[pos=tp]
\centering
\includegraphics[width=\linewidth,height=.72\textheight,keepaspectratio]{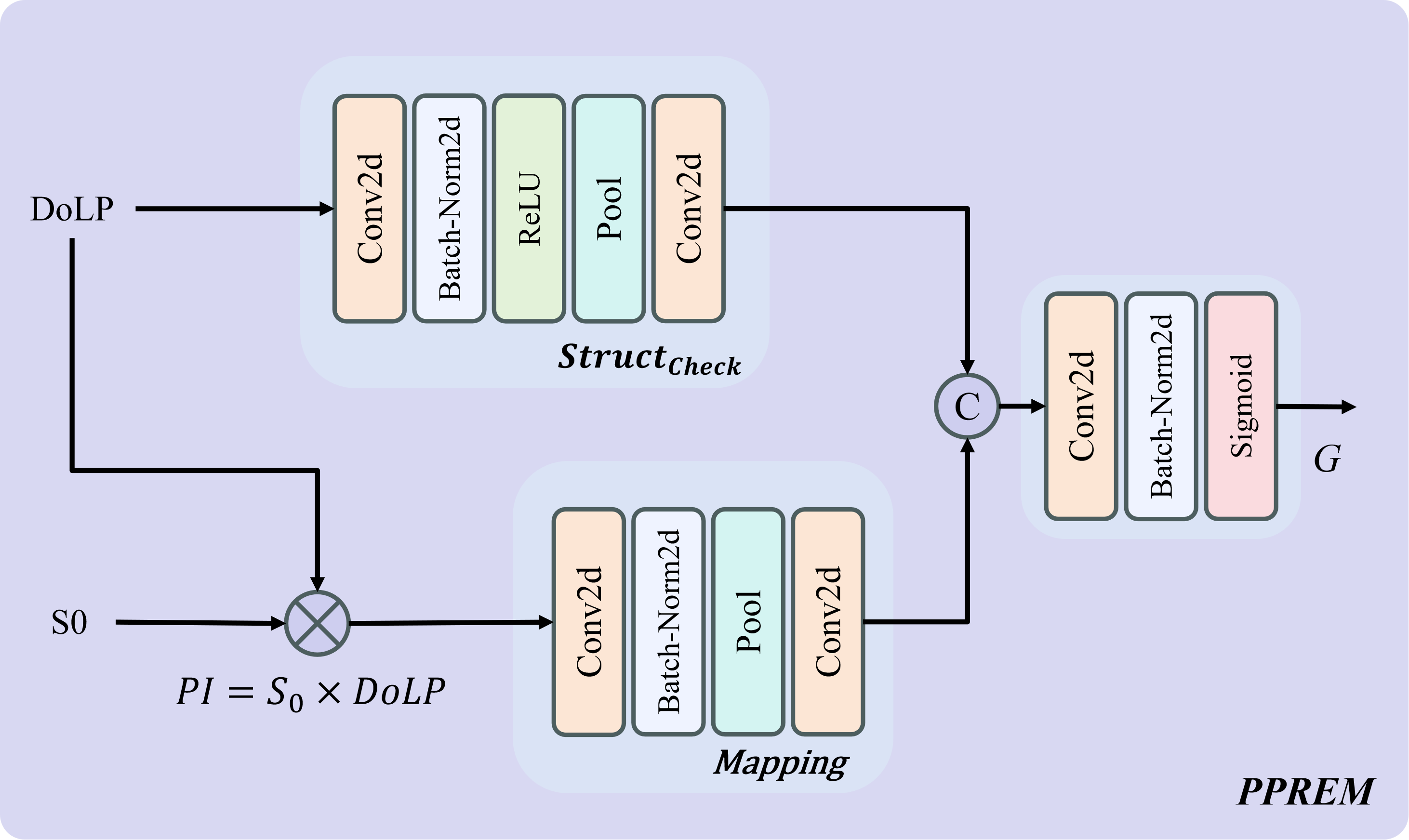}
\caption{Architecture of PPREM.}
\label{fig:7}
\end{figure}

PPREM uses physical relationships inherent in the input imagery to guide learned feature enhancement. It combines polarization information with spatial structural continuity to generate a continuous guidance map for subsequent intensity--polarization interaction. As shown in Fig.~\ref{fig:7}, it consists of a polarization response branch and a spatial structural consistency branch.
\paragraph{Polarization response branch.}
Under ideal radiometric calibration and consistent physical units, Stokes theory gives the linearly polarized intensity as
\begin{equation}\label{eq:pi}
\mathrm{PI}=\sqrt{S_1^2+S_2^2}=S_0\,\mathrm{DoLP}.
\end{equation}
Here, PI is the amplitude of the linearly polarized component determined by $S_1$ and $S_2$. Shen et al.\cite{ref24} use $I_S(x,y)=\sqrt{S_1^2(x,y)+[S_2(x,y)+S_1(x,y)]^2}$ to improve target detection. Motivated by this use of polarization-derived intensity, we use $S_0\,\mathrm{DoLP}$ to suppress polarization noise while preserving vehicle information. Elementwise interaction yields an intensity-supported polarization response:
\begin{equation}\label{eq:fp}
F_p=S_0\odot\mathrm{DoLP}.
\end{equation}

\paragraph{Spatial structural consistency branch.}
Intensity support alone can weaken polarization structures in regions with low thermal contrast. When target and background intensities are similar, filtering DoLP solely through $S_0$ may attenuate contours arising from material or geometric differences. The structural branch therefore learns local consistency directly from DoLP features, so that prior estimation accounts for both intensity support and structural continuity. A $3\times3$ convolution, average pooling, and batch normalization produce
\begin{equation}\label{eq:fs}
F_s=\operatorname{Struct}(F_{\mathrm{DoLP}}).
\end{equation}
The operator $\operatorname{Struct}(\cdot)$ aggregates local neighborhoods. Continuous edges, coherent regions, and stable geometric structures can produce consistent activations, whereas isolated random responses have limited spatial support. This reduces sensitivity to individual high-response pixels. 

\paragraph{Prior guidance generation.}
The two branch outputs are concatenated and transformed as
\begin{equation}\label{eq:fg}
F_g=\operatorname{Concat}(F_p,F_s),
\end{equation}
\begin{equation}\label{eq:guide}
G=\sigma\bigl(\operatorname{Conv}(F_g)\bigr).
\end{equation}
The sigmoid function produces continuous guidance $G$, representing relative confidence at different spatial locations. Cross-Mamba uses this guidance to regulate cross-modal feature propagation.

\subsection{Multidirectional Cross-Mamba Fusion Module}
\begin{figure}[pos=tp]
\centering
\includegraphics[width=\linewidth,height=.72\textheight,keepaspectratio]{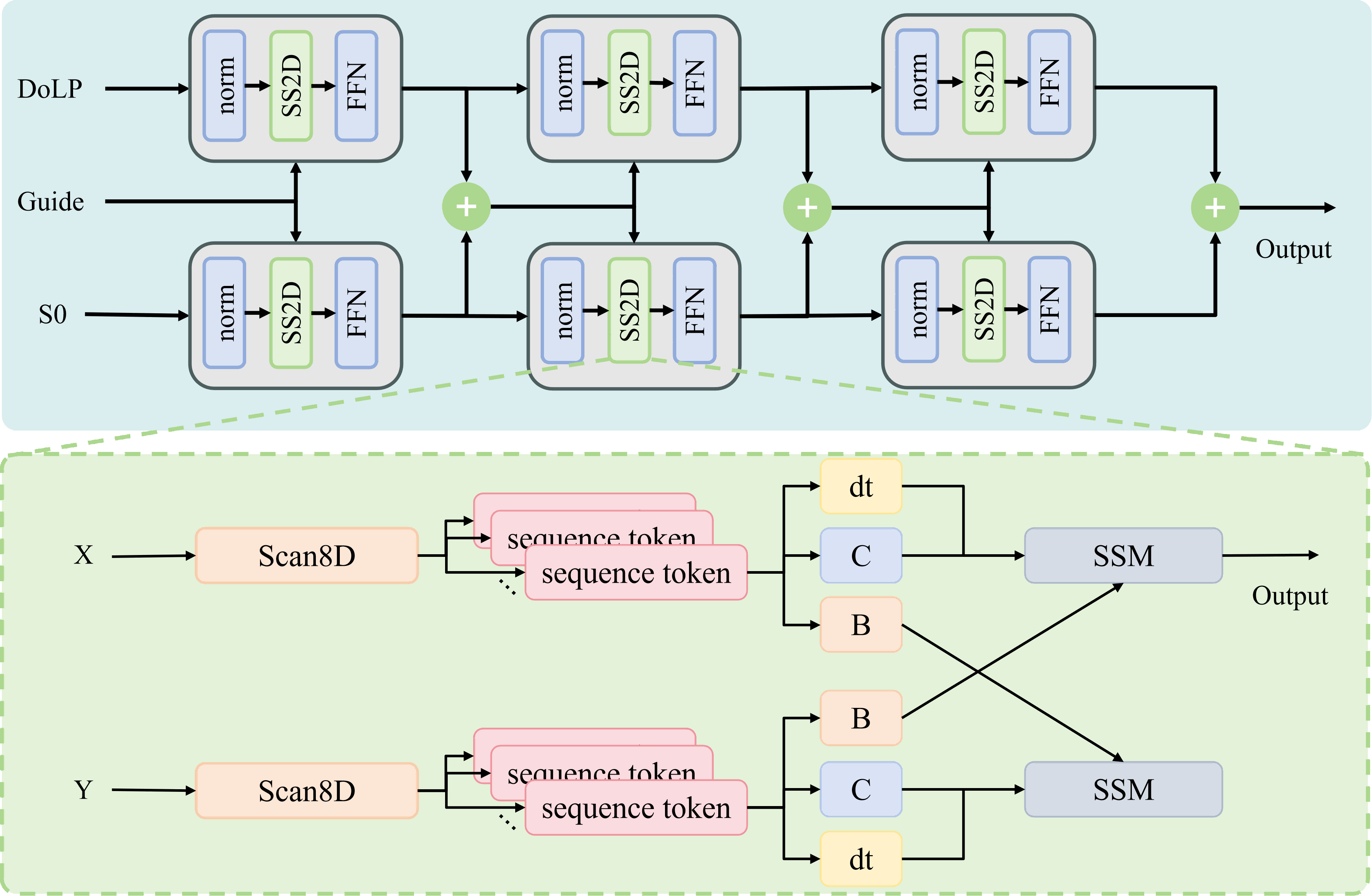}
\caption{Architecture of the Cross-Mamba fusion module.}
\label{fig:8}
\end{figure}

\begin{figure}[pos=tp]
\centering
\includegraphics[width=\linewidth,height=.72\textheight,keepaspectratio]{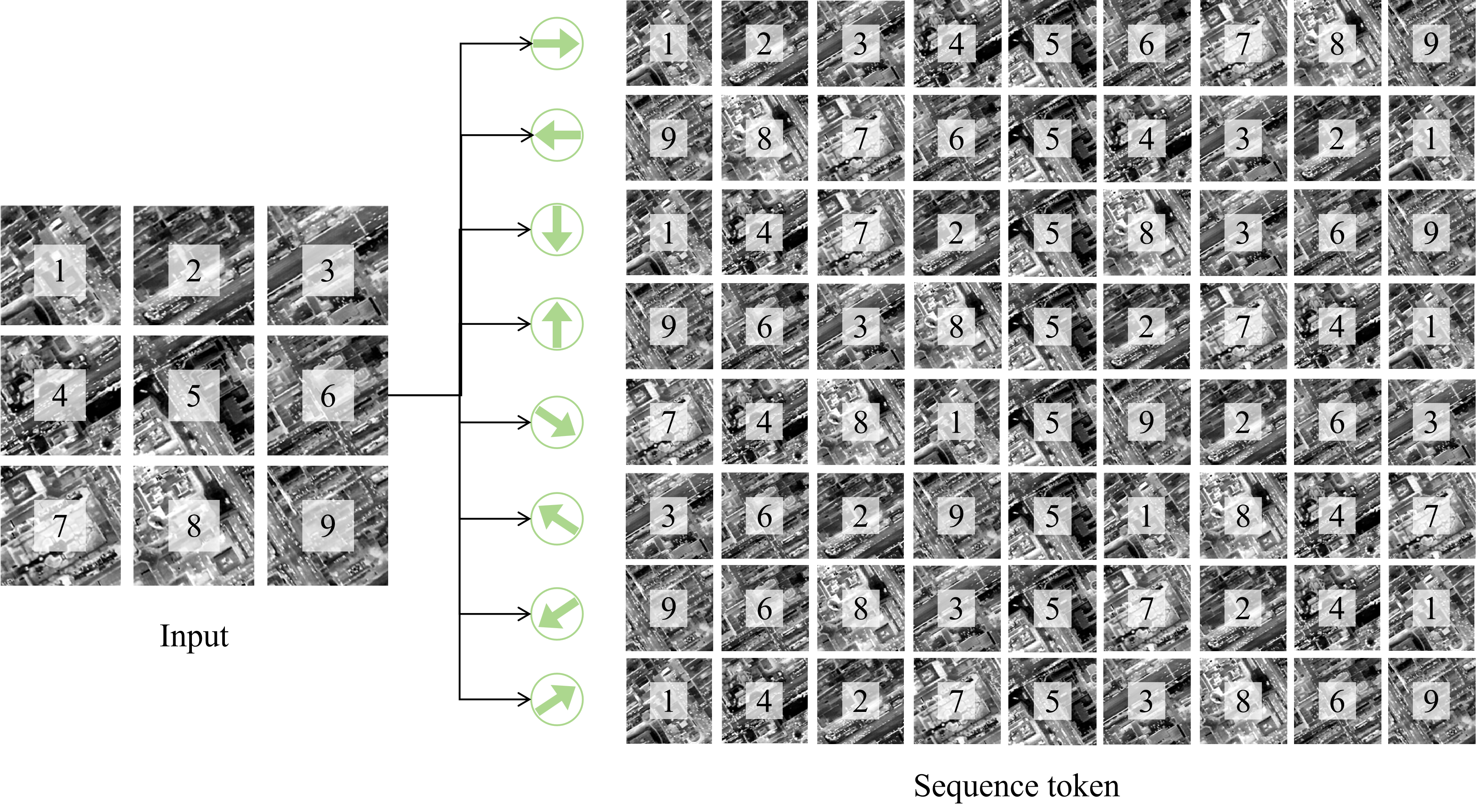}
\caption{Eight-direction scanning in Cross-Mamba.}
\label{fig:9}
\end{figure}

Building on the Mamba model studied by Cao et al.\cite{ref25}, we construct a prior-guided Cross-Mamba module with infrared intensity features $I_{S_0}$, polarization features $I_{\mathrm{DoLP}}$, and guidance $G$ as inputs. Multistage interaction progressively updates the modality representations. In the first stage (Fig.~\ref{fig:8}), two independent Mamba branches combine the modality features with $G$:
\begin{equation}
F_{S_0}^{(1)}=\mathcal{M}_{S_0}(I_{S_0},G),
\end{equation}
\begin{equation}
F_{\mathrm{DoLP}}^{(1)}=\mathcal{M}_{\mathrm{DoLP}}(I_{\mathrm{DoLP}},G).
\end{equation}
Here, $\mathcal{M}_{S_0}$ and $\mathcal{M}_{\mathrm{DoLP}}$ denote the modeling units in the two branches. Guidance provides auxiliary information for adjusting spatial responses. The enhanced features are then passed to the eight-direction scanning Mamba module for cross-modal interaction, yielding
\begin{equation}
F_f^{(1)}=F_{S_0}^{(1)}+F_{\mathrm{DoLP}}^{(1)}.
\end{equation}
The current fused feature interacts with the original modality features to obtain the next-stage representations:
\begin{equation}
F_{S_0}^{(2)}=\mathcal{M}_{S_0}^{(2)}(I_{S_0},F_f^{(1)}),
\end{equation}
\begin{equation}
F_{\mathrm{DoLP}}^{(2)}=\mathcal{M}_{\mathrm{DoLP}}^{(2)}(I_{\mathrm{DoLP}},F_f^{(1)}).
\end{equation}
The updated features again undergo eight-direction scanning. Repeating this interaction twice produces the final fused features:
\begin{equation}
F_{\mathrm{fuse}}=F_{S_0}^{(3)}+F_{\mathrm{DoLP}}^{(3)}.
\end{equation}
Roads, building boundaries, and vehicle contours are spatially continuous but need not align with the horizontal or vertical axes. We therefore incorporate both diagonal orientations into state space scanning to extend propagation paths across differently oriented structures (Fig.~\ref{fig:9}). Given $X\in\mathbb{R}^{B\times C\times H\times W}$, spatial locations are represented by an index sequence of length $L=HW$. Horizontal scanning traverses rows from top to bottom and each row from left to right. Vertical scanning traverses columns from left to right and each column from top to bottom. Main-diagonal scanning enumerates diagonal offsets from $-(H-1)$ to $W-1$ and traverses each diagonal from top left to bottom right. Anti-diagonal scanning first flips the spatial indices horizontally and applies the same diagonal enumeration, producing top-right-to-bottom-left sequences in the original coordinates.
To reduce directional bias, each of the four sequences is also reversed in its entirety, giving eight bidirectional paths across four spatial orientations. If $\pi_k$ denotes the index permutation for path $k$, the scanned sequence is
\begin{equation}
X_k=\operatorname{Gather}(\operatorname{vec}(X),\pi_k),\quad k=1,\ldots,8.
\end{equation}
The resulting tensor $X_{\mathrm{scan}}\in\mathbb{R}^{B\times8\times C\times L}$ is processed by the state space models. For each output sequence $Y_k$, the inverse permutation $\pi_k^{-1}$, obtained by applying $\operatorname{argsort}$ to the scan indices, restores the original spatial ordering. Aligned outputs are summed elementwise:
\begin{equation}
Y=\sum_{k=1}^{8}\operatorname{Gather}(Y_k,\pi_k^{-1}).
\end{equation}
Finally, $Y$ is reshaped into a two-dimensional feature map. Multidirectional information is thus aggregated by direct summation after spatial alignment.

\subsection{Loss Functions}\label{sec:loss}
The fusion network is trained without pixel-level fusion ground truth. Because infrared intensity and DoLP differ in both represented content and physical meaning, a single pixelwise constraint cannot adequately preserve scene backgrounds, local polarization structures, and stable modality representations. We combine a prior-weighted multi-window structural similarity loss, an encoder--decoder reconstruction loss, and a relative lower-bound loss for polarization saliency:
\begin{equation}\label{eq:total}
\mathcal{L}_{\mathrm{total}}=\mathcal{L}_{\mathrm{ssim}}+\lambda_{\mathrm{rec}}\mathcal{L}_{\mathrm{rec}}+\lambda_{\mathrm{psr}}\mathcal{L}_{\mathrm{psr}}.
\end{equation}
The weights are empirically set to $\lambda_{\mathrm{rec}}=0.25$ and $\lambda_{\mathrm{psr}}=1$. The SSIM term uses PPREM guidance to adapt the structural constraints from $S_0$ and DoLP. Reconstruction preserves basic modality information during feature transformation, while the polarization saliency term applies a one-sided relative penalty where the fused response falls below the reference proxy.

\subsubsection{Prior-Weighted Multi-Window Structural Similarity Loss}
Infrared intensity provides stable background and overall grayscale information. DoLP is sensitive to material- and geometry-dependent local contours but also contains unstable high responses. Fixed modality weights can either attenuate polarization details or introduce excessive noise. We therefore use PPREM guidance to adapt the structural weights spatially and average the resulting losses over several local window sizes. Given guidance channels $G_c$, with $C_g$ channels in total, we first compute
\begin{equation}
\overline{G}=\frac{1}{C_g}\sum_{c=1}^{C_g}G_c.
\end{equation}
Min--max normalization is performed independently for each sample:
\begin{equation}
M=\frac{\overline{G}-\min(\overline{G})}{\max(\overline{G})-\min(\overline{G})+\varepsilon},
\end{equation}
where $M\in[0,1]^{B\times1\times H\times W}$ and $\varepsilon=10^{-5}$ prevents division by zero. During loss computation, we apply $M\leftarrow\operatorname{sg}(M)$, where $\operatorname{sg}$ denotes stop-gradient. Thus, the guidance serves as a spatial weight without allowing the network to reduce the structural loss by directly changing that weight.

Following Li et al.\cite{ref20}, we use Gaussian windows of sizes $\mathcal{W}=\{11,9,7,5,3\}$ at the same image resolution. For a window size $w\in\mathcal{W}$, local structural similarity is
\begin{equation}
\operatorname{SSIM}_w(X,Y)=\frac{(2\mu_X\mu_Y+C_1)(2\sigma_{XY}+C_2)}{(\mu_X^2+\mu_Y^2+C_1)(\sigma_X^2+\sigma_Y^2+C_2)}.
\end{equation}
Here, $\mu_X$ and $\mu_Y$ are Gaussian-window means, $\sigma_X^2$ and $\sigma_Y^2$ are local variances, and $\sigma_{XY}$ is the covariance. We set $C_1=0.01^2$, $C_2=0.03^2$, and the Gaussian standard deviation to 1.5. The similarity maps between the fused image $I_f$ and the sources are
\begin{equation}
S_{S_0}^{w}=\operatorname{SSIM}_w(I_{S_0},I_f),
\end{equation}
\begin{equation}
S_{\mathrm{DoLP}}^{w}=\operatorname{SSIM}_w(I_{\mathrm{DoLP}},I_f).
\end{equation}
The spatially weighted loss for one window size is
\begin{equation}
\begin{split}
\mathcal{L}_{\mathrm{ssim}}^{w}=\frac{1}{N}\sum_{i=1}^{N}\bigl[&(1-M_i)(1-S_{S_0,i}^{w})\\
&+M_i(1-S_{\mathrm{DoLP},i}^{w})\bigr],
\end{split}
\end{equation}
where $i$ indexes pixels and $N=BHW$. Averaging over window sizes gives
\begin{equation}
\mathcal{L}_{\mathrm{ssim}}=\frac{1}{|\mathcal{W}|}\sum_{w\in\mathcal{W}}\mathcal{L}_{\mathrm{ssim}}^{w}.
\end{equation}
Low guidance responses emphasize the stable background and basic structures in $S_0$. High responses increase structural consistency with DoLP, helping retain polarization contours with spatial support.

\subsubsection{Encoder--Decoder Reconstruction Loss}
Following Li et al.\cite{ref20}, we also constrain the encoding and decoding processes of the dual-branch LLRR network. Let $X_{S_0}$ and $X_{\mathrm{DoLP}}$ denote the reference representations and $\widehat{X}_{S_0}$ and $\widehat{X}_{\mathrm{DoLP}}$ their reconstructions. The loss is
\begin{equation}
\begin{split}
\mathcal{L}_{\mathrm{rec}}={}&\|\widehat{X}_{S_0}-X_{S_0}\|_1\\
&+\|\widehat{X}_{\mathrm{DoLP}}-X_{\mathrm{DoLP}}\|_1.
\end{split}
\end{equation}
This auxiliary constraint retains basic modality information during feature transformation and limits degradation of representations driven solely by the fusion objective. Its weight of 0.25 maintains this regularizing role without dominating fused-image optimization.

\subsubsection{Relative Lower-Bound Loss for Polarization Saliency}
The intensity image $S_0$ represents the thermal radiation distribution, whereas DoLP describes the relative linearly polarized component. For normalized inputs, their pixelwise product defines a polarization-related response:
\begin{equation}\label{eq:proxy}
P(i)=I_{S_0}(i)I_{\mathrm{DoLP}}(i),
\end{equation}
where $i$ is a pixel location and $P$ is the polarization saliency reference map. Under ideal linear polarimetric imaging,
\begin{equation}
\mathrm{DoLP}=\frac{\sqrt{S_1^2+S_2^2}}{S_0},
\end{equation}
\begin{equation}
\mathrm{PI}=S_0\,\mathrm{DoLP}=\sqrt{S_1^2+S_2^2}.
\end{equation}
This product gives the amplitude of the linearly polarized component. Because our network inputs undergo normalization and grayscale mapping, their product is used as a polarization saliency proxy rather than a radiometrically calibrated intensity. It identifies regions supported by both intensity and polarization responses.

A conventional two-source maximum constraint drives the fused image toward the pixelwise maximum of the inputs throughout the image. Since $S_0$ and DoLP represent an intensity and a dimensionless ratio, respectively, this can impose overly strong grayscale constraints. We instead penalize only responses below the polarization saliency reference. Define the lower-bound violation as
\begin{equation}
V(i)=[P(i)-I_f(i)]_+,
\end{equation}
where
\begin{equation}
[x]_+=\max(x,0).
\end{equation}
If $I_f(i)\geq P(i)$, then $V(i)=0$ and no additional penalty is applied. If $I_f(i)<P(i)$, the loss acts to reduce attenuation of the polarization-related response. The violation mask is
\begin{equation}
M_v(i)=\mathbf{1}[V(i)>0],
\end{equation}
where $\mathbf{1}[\cdot]$ is the indicator function. The number of violating pixels is
\begin{equation}
N_v=\sum_{i=1}^{N}M_v(i),\qquad N=BHW.
\end{equation}
Absolute violations can make the loss predominantly sensitive to high-response regions. We normalize by the reference response to obtain the relative violation:
\begin{equation}
R_v(i)=\frac{V(i)}{P(i)+\varepsilon}=\frac{[P(i)-I_f(i)]_+}{P(i)+\varepsilon},
\end{equation}
with $\varepsilon=10^{-5}$. This measures the fractional response deficit, reduces domination by high-response regions, and stabilizes the denominator in low-response regions. The loss is
\begin{equation}\label{eq:psr}
\mathcal{L}_{\mathrm{psr}}=\frac{\displaystyle\sum_{i=1}^{N}M_v(i)\frac{[P(i)-I_f(i)]_+}{P(i)+\varepsilon}}{\displaystyle\sum_{i=1}^{N}M_v(i)+\varepsilon}.
\end{equation}
Substituting Eq.~\eqref{eq:proxy} gives
\begin{equation}
\mathcal{L}_{\mathrm{psr}}=\frac{\displaystyle\sum_{i=1}^{N}M_v(i)\frac{[I_{S_0}(i)I_{\mathrm{DoLP}}(i)-I_f(i)]_+}{I_{S_0}(i)I_{\mathrm{DoLP}}(i)+\varepsilon}}{\displaystyle\sum_{i=1}^{N}M_v(i)+\varepsilon},
\end{equation}
where
\begin{equation}
M_v(i)=\mathbf{1}[I_f(i)<I_{S_0}(i)I_{\mathrm{DoLP}}(i)].
\end{equation}
When no pixels violate the bound, the numerator and loss are zero. The constraint raises fused responses below $P$ and leaves responses already satisfying the bound unpenalized. Thus, $P$ serves as a local response floor to reduce attenuation of target contours carrying discriminative polarization information.

\section{Experiments}\label{sec:experiments}
\subsection{Experimental Settings}
Training uses $256\times256$ image patches to reduce memory use and diversify local spatial configurations. AIR-LWIR-Pol is split into 1,515 training pairs and 386 test pairs. All models are implemented using PyTorch and CUDA and trained on an NVIDIA RTX 5880 GPU. We use Adam with an initial learning rate of $10^{-4}$ and cosine annealing, a batch size of 4, 200 epochs, and a fixed random seed of 42. Test images are processed at their original resolution of $1280\times1024$.

We compare PolMamba with RFN-Nest\cite{ref1}, PIPFNet\cite{ref20}, GANMcC\cite{ref28}, GAN-FM\cite{ref29}, TIPFNet\cite{ref30}, CMTFusion\cite{ref31}, MSPFusion\cite{ref32}, and SFMFusion\cite{ref33}. Quantitative evaluation and visual comparisons cover grayscale statistics, source structure and contrast preservation, and no-reference perceptual quality. Six metrics are used: entropy (EN)\cite{ref34}, standard deviation (SD), SSIM\cite{ref36}, Chen--Blum fusion quality (QCB)\cite{ref39}, NIMA\cite{ref38}, and CEIQ\cite{ref39}.

EN measures uncertainty in the grayscale distribution, while SD measures overall grayscale dispersion. Together, they describe statistical information content and overall contrast. SSIM measures structural similarity between fused and source images. QCB evaluates source local-contrast preservation using contrast sensitivity and saliency weighting derived from human visual perception. NIMA and CEIQ provide no-reference scores through a learned assessment model and contrast-related quality assessment, respectively, complementing the source-based metrics.

\subsection{Comparison with Representative Methods in Remote Sensing Scenes}
\subsubsection{Quantitative Analysis}
Table~\ref{tab:1} shows that PolMamba achieves the highest EN, SD, SSIM, and NIMA, with values of 7.5711, 61.4475, 0.5337, and 5.1722, respectively, and ranks second in QCB and CEIQ. These results show balanced performance across grayscale statistics, structural similarity, local contrast preservation, and perceptual quality on the AIR-LWIR-Pol test set.

\begin{table*}[pos=tp]
\centering
\caption{Quantitative comparison of image fusion methods on the AIR-LWIR-Pol test set. Best values are shown in bold, and second-best values are underlined.}
\label{tab:1}
\small
\setlength{\tabcolsep}{3pt}
\renewcommand{\arraystretch}{1.12}
\begin{tabular*}{\textwidth}{@{\extracolsep{\fill}}lllllll@{}}
\toprule
Method & EN($\uparrow$) & SD($\uparrow$) & SSIM($\uparrow$) & QCB($\uparrow$) & NIMA($\uparrow$) & CEIQ($\uparrow$) \\
\midrule
RFN-Nest & 7.0185 & 36.9193 & 0.4588 & 0.3820 & 5.0389 & 3.1367 \\
PIPFNet & 7.4425 & \underline{59.0711} & 0.4982 & 0.4303 & \underline{5.1638} & 3.3738 \\
GANMcC & 5.6743 & 33.7843 & 0.2944 & 0.2403 & 4.9865 & 2.2051 \\
GAN-FM & 7.4551 & 49.3419 & 0.4804 & 0.4260 & 4.6279 & 1.5062 \\
TIPFNet & 6.9954 & 37.0115 & 0.5039 & \textbf{0.5040} & 4.8244 & 3.0595 \\
CMTFusion & 7.2701 & 42.1531 & 0.4903 & 0.3994 & 4.9095 & 3.2760 \\
MSPFusion & 7.3044 & 42.7936 & \underline{0.5278} & 0.3927 & 4.9723 & 3.3125 \\
SFMFusion & \underline{7.5461} & 55.6978 & -0.2505 & 0.3381 & 4.2244 & \textbf{3.4771} \\
Ours & \textbf{7.5711} & \textbf{61.4475} & \textbf{0.5337} & \underline{0.4335} & \textbf{5.1722} & \underline{3.4235} \\
\bottomrule
\end{tabular*}
\end{table*}

For grayscale statistics, PolMamba achieves an EN of 7.5711, exceeding SFMFusion at 7.5461 and GAN-FM at 7.4551. Its SD of 61.4475 exceeds PIPFNet at 59.0711 and SFMFusion at 55.6978, indicating relatively high grayscale entropy and dispersion.

For structural similarity, PolMamba achieves an SSIM of 0.5337, compared with 0.5278 for MSPFusion and 0.5039 for TIPFNet. This supports structural information preservation under the source references and SSIM implementation used here. SFMFusion has a negative average SSIM on this dataset: in some local regions, its fused images exhibit brightness variations opposite to those in DoLP, producing negative structural similarity contributions and reducing the aggregate score.

For perceptually weighted source-contrast preservation, TIPFNet achieves the highest QCB of 0.5040, followed by PolMamba at 0.4335. QCB measures the retention of source local-contrast information through human visual contrast sensitivity. PolMamba therefore outperforms most compared methods on this dimension, while TIPFNet retains the highest score.

For no-reference quality, PolMamba obtains a NIMA score of 5.1722, exceeding PIPFNet at 5.1638. Its CEIQ of 3.4235 ranks below SFMFusion at 3.4771 and above PIPFNet at 3.3738. With four first-place and two second-place results across six metrics, PolMamba combines strong structural preservation with favorable grayscale and perceptual quality scores.

\subsubsection{Qualitative Analysis}
\begin{figure*}[pos=tp]
\centering
\includegraphics[width=\linewidth,height=.82\textheight,keepaspectratio]{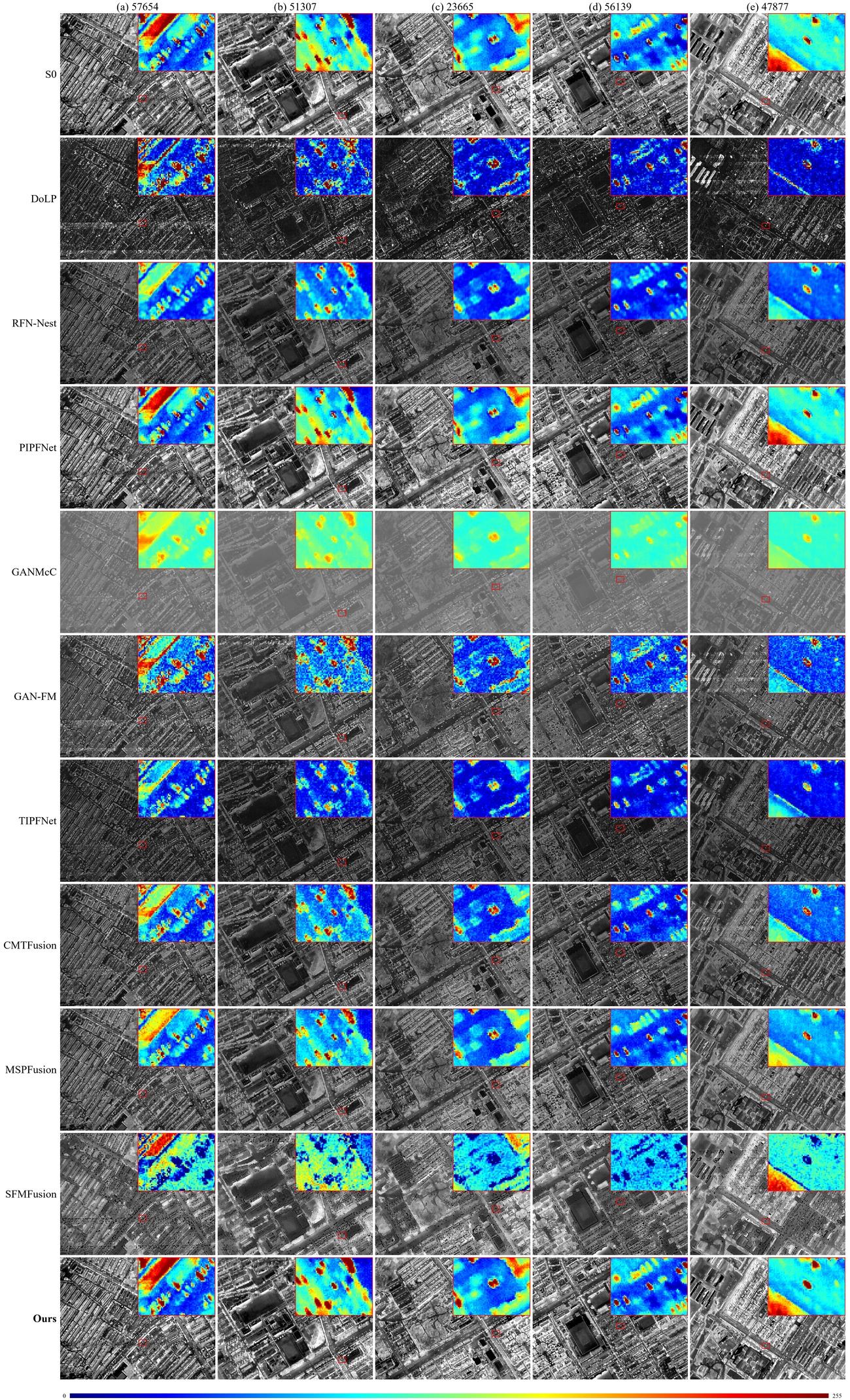}
\caption{Qualitative comparison of vehicle-region fusion results in urban scenes from AIR-LWIR-Pol.}
\label{fig:10}
\end{figure*}

We examine vehicle regions to compare scene-structure preservation and local target information. Figure~\ref{fig:10} presents the sources, fused images, and enlarged red-box regions for five scenes, allowing comparison of vehicles and their immediate backgrounds.

The $S_0$ images preserve roads and buildings, but some vehicles have limited grayscale contrast against their surroundings, particularly in Fig.~\ref{fig:10}(a). DoLP provides more prominent vehicle responses, accompanied by granular textures in some regions. PIPFNet preserves an appearance close to $S_0$, but provides limited enhancement for some weak vehicle targets. RFN-Nest, TIPFNet, CMTFusion, and MSPFusion highlight some vehicles, with different levels of background grayscale and texture preservation. GAN-FM retains conspicuous granular textures in some enlarged regions, GANMcC produces relatively low overall contrast, and SFMFusion exhibits strong grayscale variations in some regions that affect the natural appearance of targets and backgrounds.

PolMamba preserves the main road and building structures while producing clearer local vehicle responses. Vehicles are better separated from their surroundings in Fig.~\ref{fig:10}(a) and (b). In Fig.~\ref{fig:10}(c)--(e), vehicles remain discernible, adjacent backgrounds retain relatively continuous grayscale transitions, and granular textures are less pronounced than in some competing results. These examples demonstrate how polarization information complements vehicle features in intensity imagery while preserving the scene background.

\subsection{Generalization Experiments}
The method is also effective on natural scenes in the LDDRS dataset.

\subsection{Ablation Studies}
We evaluate network modules and loss terms using the same data split, training hyperparameters, and random seed, changing only the component under investigation.

\subsubsection{Module Ablation}


To evaluate the roles of PPREM and Cross-Mamba, we remove each module in turn while keeping the data split, training strategy, and evaluation settings unchanged. As shown in Table~\ref{tab:2}, the complete model achieves the highest SSIM (0.5337) and NIMA (5.1722), demonstrating its advantages in structural similarity and perceptual quality assessment under the current settings.

Without PPREM, SSIM and NIMA reach 0.5090 and 5.1634, respectively. The SSIM difference of 0.0247 supports the module's contribution to structural preservation. EN, SD, QCB, and CEIQ increase in this variant, reflecting different outcomes across the evaluation criteria. Without Cross-Mamba, five of the six metrics are lower than those of the complete model; SD is the exception. In particular, SSIM and QCB differ by 0.0199 and 0.0153, respectively, supporting the module's contribution to preserving source-image structure and local contrast information.

Both ablations yield lower SSIM and NIMA scores than the complete model, providing experimental support for the roles of the two modules. PPREM's contribution is most evident in structural similarity, while Cross-Mamba contributes across several fusion quality metrics.

\begin{figure}[pos=tp]
\centering
\includegraphics[width=\linewidth,height=.72\textheight,keepaspectratio]{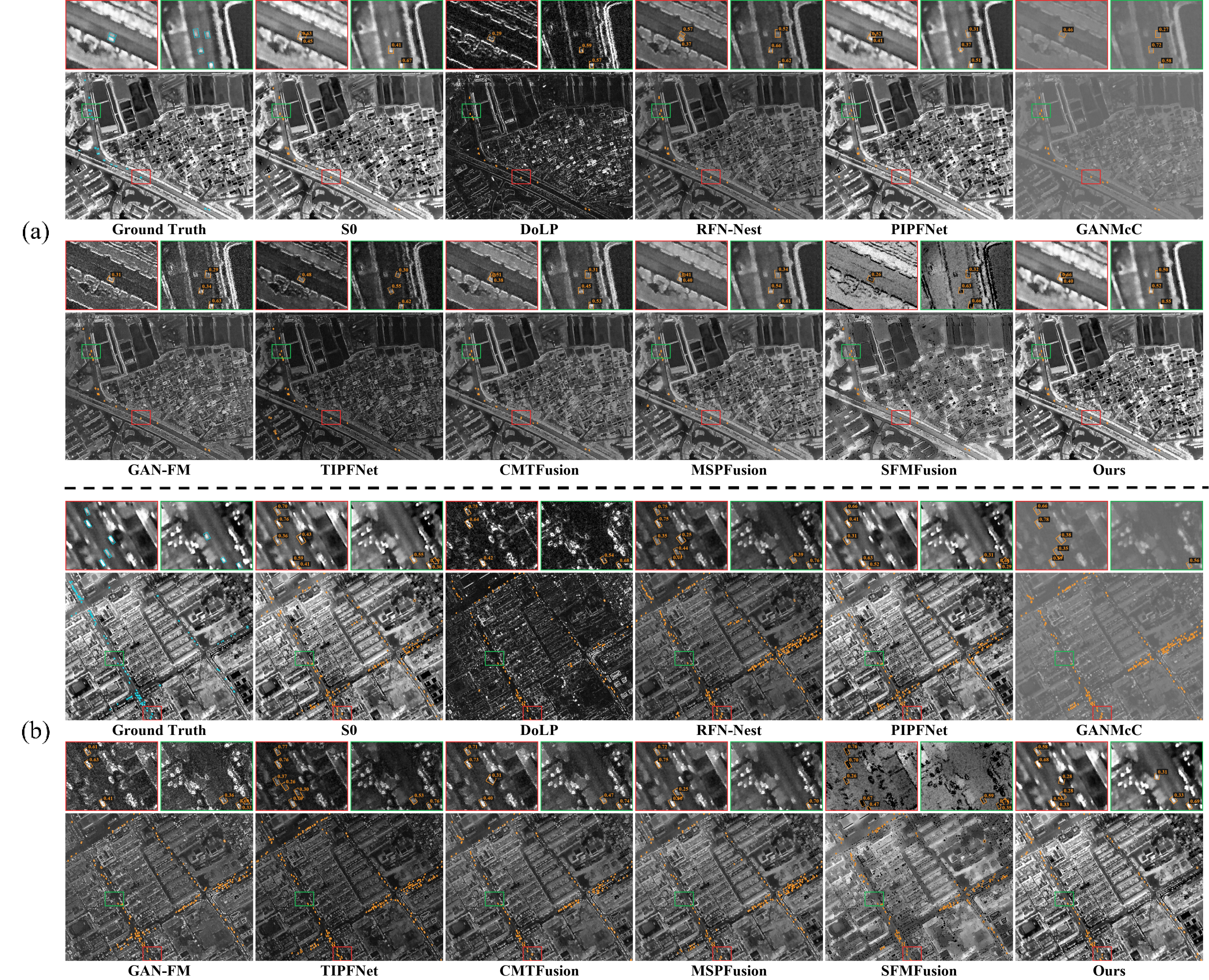}
\caption{Qualitative comparison of the module ablation configurations.}
\label{fig:11}
\end{figure}

Figure~\ref{fig:11} compares the fusion results without PPREM, without Cross-Mamba, and with the complete model. From top to bottom, the rows show grayscale images, full-image pseudocolor views, and enlarged vehicle regions. All configurations use the same pseudocolor mapping range, with color representing the 8-bit grayscale values of the fused images. The three configurations preserve the main scene structures, including roads, buildings, and vegetation; their differences are concentrated in the local grayscale representation of vehicles and their surroundings.

In the enlarged regions, removing PPREM weakens the warm-color responses of some vehicles relative to the complete model and changes their grayscale separation from the surrounding road. In the complete model, the corresponding vehicles exhibit more prominent high-grayscale responses, while the road layout and nearby building contours remain visible. Removing Cross-Mamba produces stronger high-grayscale responses in some vehicles and surrounding ground features, together with more pronounced local intensity variations. The complete model maintains discernible vehicles with more moderate background responses.

Considered alongside the quantitative results, these visualizations show that the complete model preserves both vehicle responses and background structure in the displayed scene. PPREM guides the spatial allocation of target-region responses, while Cross-Mamba contributes to cross-modal feature integration. Together, the two modules form the fusion pathway of the complete model.

\begin{table}[pos=tp]
	\centering
	\caption{Results of the module ablation study. Best values are shown in bold, and second-best values are underlined.}
	\label{tab:2}
	\resizebox{\columnwidth}{!}{%
		\begin{tabular}{@{}lllllll@{}}
			\toprule
			Method & EN($\uparrow$) & SD($\uparrow$) & SSIM($\uparrow$) & QCB($\uparrow$) & NIMA($\uparrow$) & CEIQ($\uparrow$) \\
			\midrule
			Ours & \underline{7.5711} & 61.4475 & \textbf{0.5337} & \underline{0.4335} & \textbf{5.1722} & \underline{3.4235} \\
			w/o PPREM & \textbf{7.5758} & \underline{61.7798} & 0.5090 & \textbf{0.4401} & \underline{5.1634} & \textbf{3.4322} \\
			w/o Cross-Mamba & 7.5674 & \textbf{63.7364} & \underline{0.5138} & 0.4182 & 5.1486 & 3.4170 \\
			\bottomrule
		\end{tabular}%
	}
\end{table}

\subsubsection{Loss Function Ablation}


To assess the contribution of each loss term, we separately remove the structural similarity loss $\mathcal{L}_{\mathrm{ssim}}$, the reconstruction loss $\mathcal{L}_{\mathrm{rec}}$, and the polarization saliency preservation loss $\mathcal{L}_{\mathrm{psr}}$ under the default weight settings. As shown in Table~\ref{tab:3}, the complete objective achieves the highest EN, SD, and NIMA scores, at 7.5711, 61.4475, and 5.1722, respectively.

The structural similarity constraint has the most pronounced effect. Without $\mathcal{L}_{\mathrm{ssim}}$, all six metrics attain their lowest values among the evaluated configurations. SSIM and QCB change from 0.5337 and 0.4335 to 0.3117 and 0.2577, respectively, while CEIQ changes from 3.4235 to 2.4188. The consistent changes across structural, contrast-related, and quality metrics support the importance of this constraint beyond its direct relationship with the SSIM evaluation metric.

Without $\mathcal{L}_{\mathrm{rec}}$, EN, SD, SSIM, QCB, and NIMA are slightly lower than those of the complete configuration, whereas CEIQ increases to 3.4295. The differences in SSIM and QCB are 0.0018 and 0.0028, respectively. The small overall changes indicate that reconstruction loss serves mainly as an auxiliary constraint, complementing the preservation of source-image structure and contrast information.

Without $\mathcal{L}_{\mathrm{psr}}$, EN, SD, and NIMA are 7.5490, 59.8387, and 5.1648, respectively, while SSIM, QCB, and CEIQ increase to 0.5386, 0.4475, and 3.4315. Thus, retaining this constraint favors grayscale statistics and the NIMA score under the current settings, while its optimization emphasis differs from that of the structural similarity and contrast-related metrics.

These results support a joint objective built on structural similarity and complemented by reconstruction and polarization saliency preservation constraints. The complete configuration achieves the highest EN, SD, and NIMA scores, reflecting a balance among grayscale statistics, structural preservation, and perceptual quality assessment.

\begin{figure}[pos=tp]
\centering
\includegraphics[width=\linewidth,height=.72\textheight,keepaspectratio]{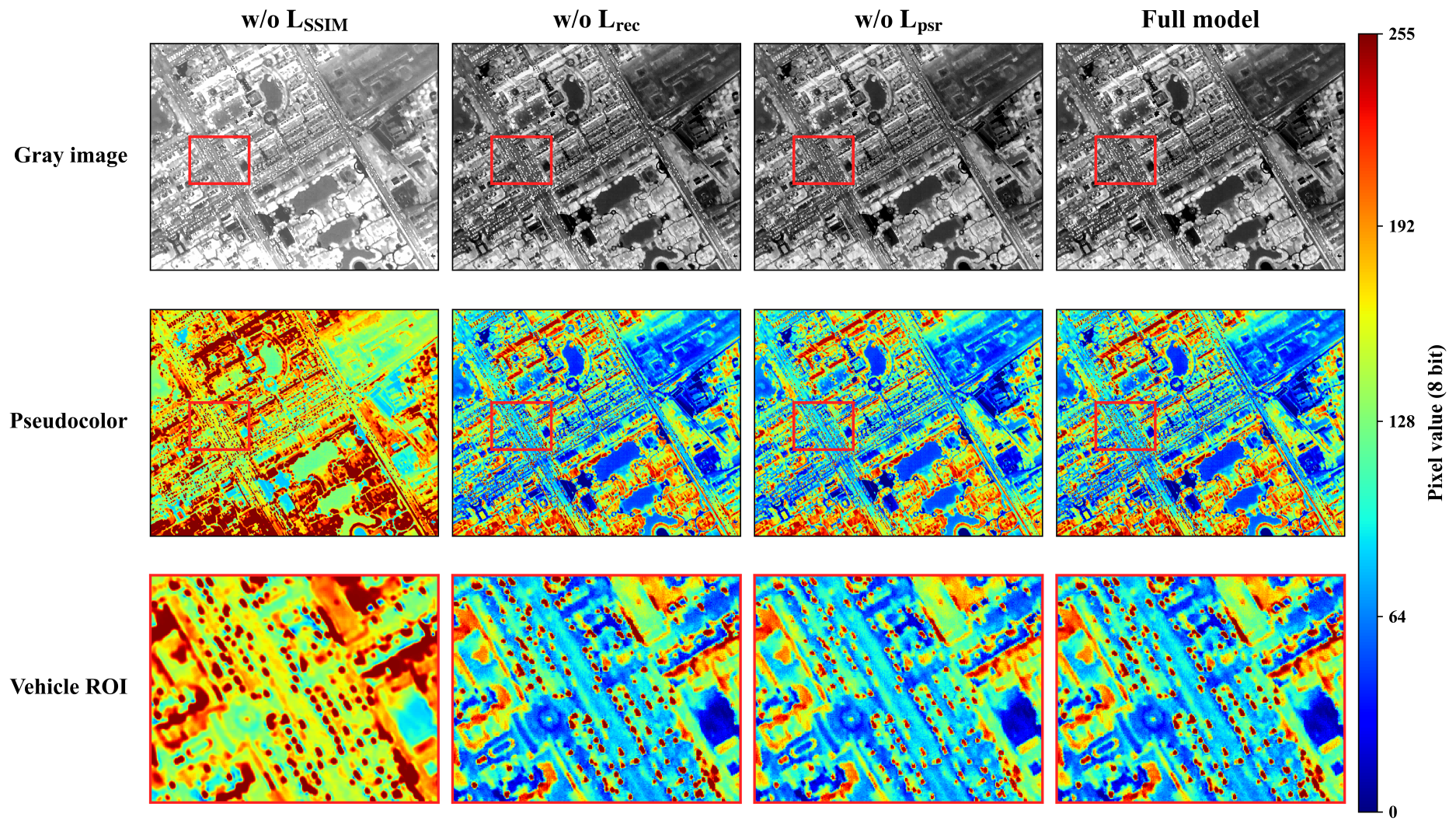}
\caption{Qualitative comparison of the loss ablation configurations.}
\label{fig:12}
\end{figure}

Figure~\ref{fig:12} presents the grayscale images, full-image pseudocolor views, and enlarged vehicle regions under different loss configurations. A common color scale represents the 8-bit grayscale values. The complete model preserves scene structures such as roads and buildings while producing concentrated high-grayscale responses at vehicle locations, making the vehicles distinguishable from their surroundings.

Without the structural similarity loss $\mathcal{L}_{\mathrm{ssim}}$, the image appears brighter overall, warm-color regions expand, vehicle-edge transitions become smoother, and some background textures are reduced. The complete model retains richer grayscale variation and more compact vehicle responses. These visual differences are consistent with the changes in SSIM and QCB and support the role of this constraint in preserving structure and local contrast.

Without the reconstruction loss $\mathcal{L}_{\mathrm{rec}}$, the overall scene layout remains close to that of the complete model. Differences are mainly visible in local textures and grayscale representation, consistent with the auxiliary information-preserving role of this loss.

Without the polarization saliency preservation loss $\mathcal{L}_{\mathrm{psr}}$, the main scene structures remain visible, but the high-grayscale responses of some vehicles weaken. The corresponding targets are more prominent in the complete model, consistent with the intended role of this constraint in reducing response attenuation in polarization-salient regions.

In the displayed scene, the complete objective preserves both background structure and local vehicle responses. These visual observations complement the quantitative loss ablation results.

\begin{table}[pos=tp]
	\centering
	\caption{Results of the loss function ablation study. Best values are shown in bold, and second-best values are underlined.}
	\label{tab:3}
	\resizebox{\columnwidth}{!}{%
		\begin{tabular}{@{}lllllll@{}}
			\toprule
			Method & EN($\uparrow$) & SD($\uparrow$) & SSIM($\uparrow$) & QCB($\uparrow$) & NIMA($\uparrow$) & CEIQ($\uparrow$) \\
			\midrule
			Ours
			& \textbf{7.5711} & \textbf{61.4475} & \underline{0.5337}
			& \underline{0.4335} & \textbf{5.1722} & 3.4235 \\
			w/o $\mathcal{L}_{\mathrm{ssim}}$
			& 6.6011 & 54.4312 & 0.3117 & 0.2577 & 4.8114 & 2.4188 \\
			w/o $\mathcal{L}_{\mathrm{rec}}$
			& \underline{7.5705} & \underline{60.7301} & 0.5319
			& 0.4307 & \underline{5.1702} & \underline{3.4295} \\
			w/o $\mathcal{L}_{\mathrm{psr}}$
			& 7.5490 & 59.8387 & \textbf{0.5386}
			& \textbf{0.4475} & 5.1648 & \textbf{3.4315} \\
			\bottomrule
		\end{tabular}%
	}
\end{table}

\section{Discussion}\label{sec:discussion}
\subsection{Sensitivity to Loss Function Weights}

\subsection{Analysis of the Two Branches in PPREM}

\begin{figure*}[pos=tp]
	\centering
	\includegraphics[
	width=\textwidth,
	height=0.72\textheight,
	keepaspectratio
	]{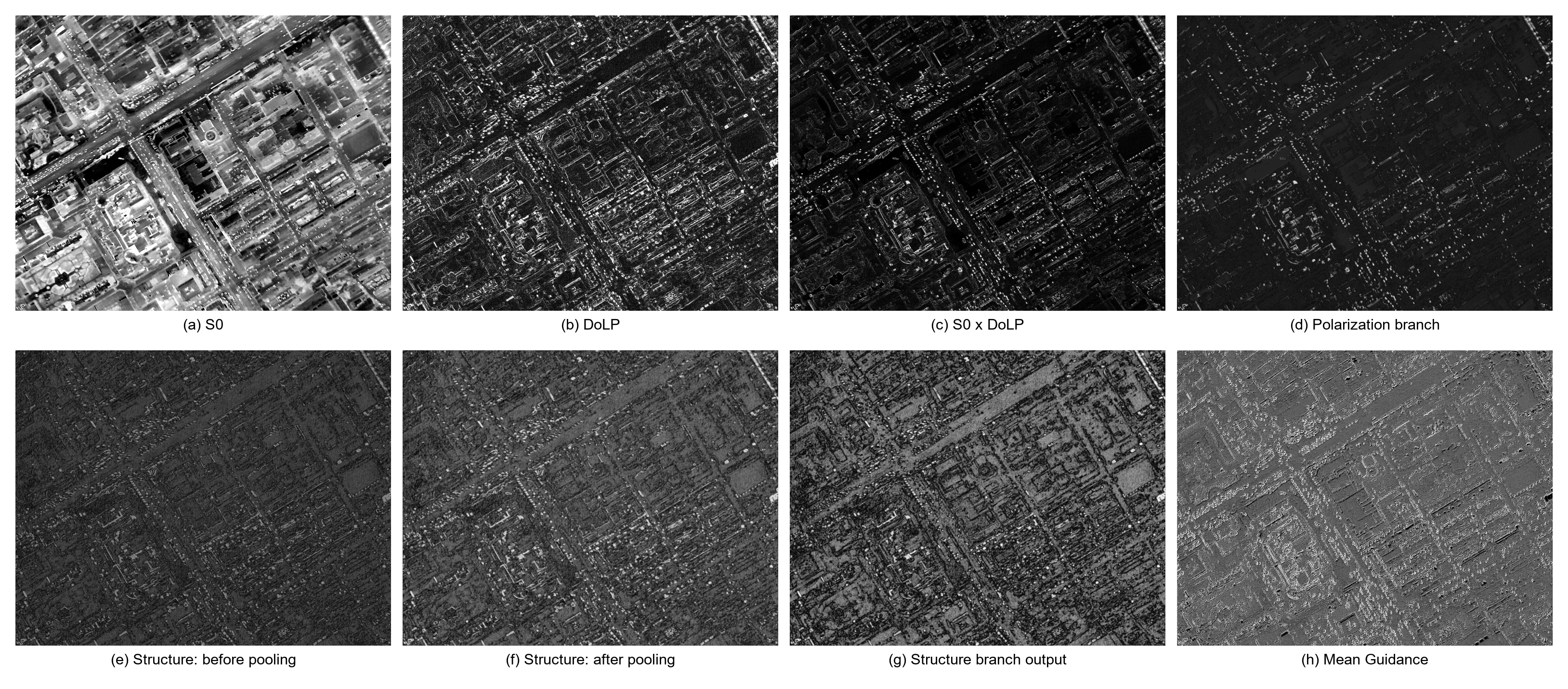}
	\caption{Intermediate outputs of PPREM:
		(a) $S_0$;
		(b) DoLP;
		(c) $S_0\odot\mathrm{DoLP}$;
		(d) polarization response branch;
		(e) structural features before pooling;
		(f) structural features after pooling;
		(g) spatial structural consistency branch output;
		(h) channel-averaged guidance map.}
	\label{fig:13}
\end{figure*}

PPREM constructs physics-guided features through two complementary extraction paths: an intensity-modulated polarization response path and a path that captures spatial structure in DoLP. Figure~\ref{fig:13} shows intermediate results from the two branches for the same scene. The branches use the pixelwise product of $S_0$ and DoLP and the DoLP image, respectively, to represent polarization information from different perspectives. Their features are combined to generate multichannel guidance weights.

The polarization response branch constructs a joint prior using the pixelwise product $S_0\odot\mathrm{DoLP}$. As shown in Fig.~\ref{fig:13}(b) and (c), this product retains some prominent polarization responses while modulating their spatial distribution by $S_0$. A subsequent convolutional mapping transforms the joint response into multichannel features. The bright points and fine lines in Fig.~\ref{fig:13}(d) are concentrated around selected scene structures, indicating a spatially selective response. Because the product is computed from normalized network inputs, it represents joint intensity and polarization information at the input scale.

The spatial structural consistency branch extracts features directly from DoLP, complementing the joint prior with neighborhood information. Convolution and nonlinear mapping produce local responses, which are then aggregated by $3\times3$ average pooling. Figures~\ref{fig:13}(e) and (f) show the features before and after pooling, respectively. The branch output in Fig.~\ref{fig:13}(g) represents the spatial distribution of roads, buildings, and surrounding textures. Since this path operates directly on DoLP, its structural representation is not constrained by the pixelwise product with $S_0$, making it complementary to the polarization response branch.

The eight-channel features from each branch are concatenated and passed through convolution, batch normalization, and a sigmoid function to produce 32-channel guidance weights. Figure~\ref{fig:13}(h) shows the channel-averaged guidance map; display stretching makes its scene-related spatial variations visible. The spatial means of individual channels range from approximately 0.390 to 0.623, reflecting differences among their weights. PPREM thus combines selective responses from the joint prior with DoLP-based structural features to guide subsequent feature modulation.

\subsection{Analysis of the Number of Mamba Blocks in Cross-Mamba}

\subsection{Vehicle Detection}\label{sec:detection}


To evaluate the utility of fused images for downstream vehicle detection, we train YOLO11s-OBB detectors using the source images and the fused images produced by each method. All configurations use the same network architecture, data split, and training hyperparameters, and are initialized with the official pretrained weights \texttt{yolo11s-obb.pt}.

As shown in Table~\ref{tab:5}, our method achieves the highest precision, $\mathrm{mAP}_{50}$, and F1 scores of 0.607, 0.574, and 0.561, respectively. Its $\mathrm{mAP}_{50\text{--}95}$ of 0.303 ranks second. All five detection metrics improve over DoLP input. Relative to $S_0$ input, precision and $\mathrm{mAP}_{50}$ increase by 0.056 and 0.015, respectively, demonstrating the utility of the fused images for vehicle detection under the current settings.

Compared with PIPFNet, our method improves precision by 0.043 and $\mathrm{mAP}_{50}$ by 0.005. Their F1 scores are close, at 0.561 and 0.560, respectively. PIPFNet achieves the highest recall and $\mathrm{mAP}_{50\text{--}95}$, at 0.557 and 0.317, compared with 0.521 and 0.303 for our method. The two methods thus emphasize different aspects of precision, recall, and detection performance across IoU thresholds.

Overall, our method ranks first on three of the five detection metrics and second on one. Its advantages in precision and $\mathrm{mAP}_{50}$ provide experimental support for using the fused images in oriented vehicle detection.

\begin{figure*}[pos=tp]
\centering
\includegraphics[width=\linewidth,height=.72\textheight,keepaspectratio]{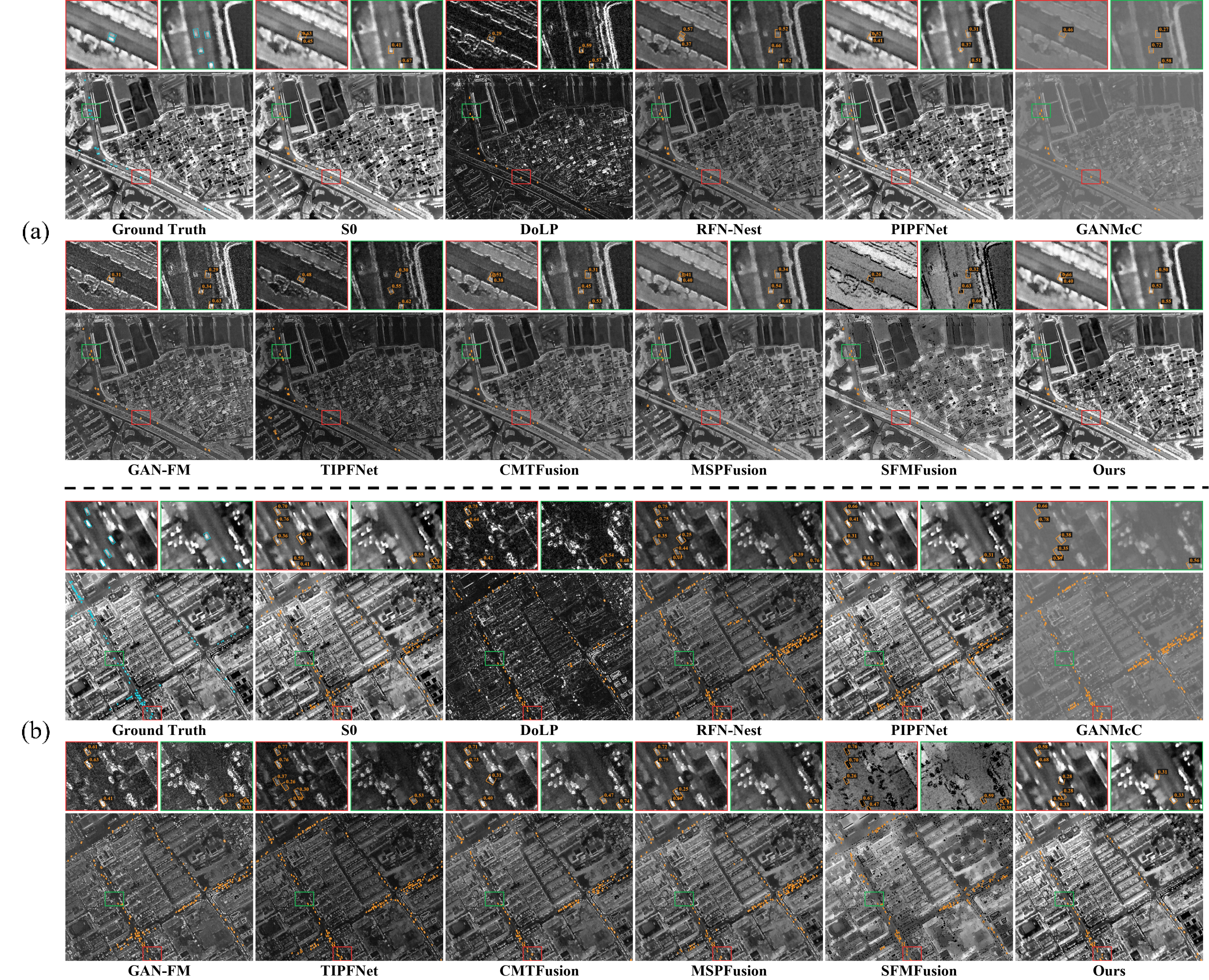}
\caption{Qualitative comparison of oriented vehicle detection using different source and fused images.}
\label{fig:14}
\end{figure*}

Figure~\ref{fig:14} shows oriented vehicle detection results in two complex road scenes. Cyan boxes denote ground-truth annotations, and orange boxes denote predictions. All methods use a confidence threshold of 0.25 and an oriented-polygon intersection-over-union (IoU) matching threshold of 0.50, allowing comparison of small-vehicle detection and false positives.

In scene (a), our method correctly detects two and three vehicles in the red and green enlarged regions, respectively. Across the full image, it correctly detects 16 of the 20 annotated vehicles with one false positive. Its precision, recall, and F1 are 94.12\%, 80.00\%, and 86.49\%, respectively, the highest among the compared configurations. GAN-FM and $S_0$ each correctly detect 15 vehicles but produce two and three false positives, respectively. These results show that our fused images support both vehicle detection and false-positive control in this scene.

In the more densely populated scene (b), our method correctly detects four of the five vehicles in the red region and all three vehicles in the green region. Across the full image, it correctly detects 50 of the 93 annotated vehicles with 24 false positives. Its precision of 67.57\% and F1 of 59.88\% exceed those of the other methods. RFN-Nest correctly detects 57 vehicles but produces 164 false positives, yielding precision and F1 of 25.79\% and 36.31\%, respectively. This comparison highlights the false-positive control achieved by our method in the dense scene.

Together, the enlarged regions and full-image statistics show that the fused images support vehicle recognition in complex backgrounds while balancing target detection and prediction reliability.

\begin{table}[pos=tp]
	\centering
	\caption{Quantitative comparison of source and fused images for downstream vehicle detection. Best values are shown in bold, and second-best values are underlined.}
	\label{tab:5}
	\resizebox{\columnwidth}{!}{%
		\begin{tabular}{@{}llllll@{}}
			\toprule
			Method
			& \shortstack{Precision($\uparrow$)}
			& \shortstack{Recall($\uparrow$)}
			& \shortstack{$\mathrm{mAP}_{50}$($\uparrow$)}
			& \shortstack{$\mathrm{mAP}_{50\text{--}95}$($\uparrow$)}
			& \shortstack{F1($\uparrow$)} \\
			\midrule
			$S_0$       & 0.551 & \underline{0.548} & 0.559 & 0.302 & 0.549 \\
			DoLP        & 0.498 & 0.454 & 0.441 & 0.209 & 0.475 \\
			RFN-Nest    & 0.511 & 0.493 & 0.485 & 0.261 & 0.502 \\
			PIPFNet     & \underline{0.564} & \textbf{0.557} & \underline{0.569} & \textbf{0.317} & \underline{0.560} \\
			GANMcC      & 0.469 & 0.490 & 0.474 & 0.247 & 0.479 \\
			GAN-FM      & 0.447 & 0.524 & 0.460 & 0.226 & 0.482 \\
			TIPFNet     & 0.497 & 0.499 & 0.476 & 0.250 & 0.498 \\
			CMTFusion   & 0.505 & 0.534 & 0.501 & 0.253 & 0.519 \\
			MSPFusion   & 0.506 & 0.512 & 0.481 & 0.259 & 0.509 \\
			SFMFusion   & 0.535 & 0.415 & 0.441 & 0.216 & 0.467 \\
			Ours        & \textbf{0.607} & 0.521 & \textbf{0.574} & \underline{0.303} & \textbf{0.561} \\
			\bottomrule
		\end{tabular}%
	}
\end{table}

\subsection{Computational Efficiency}
We assess model complexity and efficiency using the number of parameters, floating-point operations (FLOPs), and average inference time. The efficiency experiment uses AIR-LWIR-Pol images of $1259\times990$ pixels. As shown in Table~\ref{tab:6}, PolMamba has 1.6479 million parameters, fewer than RFN-Nest, GANMcC, GAN-FM, and TIPFNet, but more than PIPFNet, CMTFusion, MSPFusion, and SFMFusion. Its parameter count is therefore intermediate among the compared methods.

\begin{table}[pos=tp]
	\centering
	\caption{Comparison of model complexity and computational efficiency. Best values are shown in bold, and second-best values are underlined.}
	\label{tab:6}
	\small
	\setlength{\tabcolsep}{3pt}
	\renewcommand{\arraystretch}{1.12}
	\begin{tabular*}{\columnwidth}{@{\extracolsep{\fill}}llll@{}}
		\toprule
		Method & Params(M)$\downarrow$ & FLOPs(G)$\downarrow$ & Time(s)$\downarrow$ \\
		\midrule
		RFN-Nest  & 7.5242 & 4221.8378 & 0.2070 \\
		PIPFNet   & \textbf{0.0346} & \textbf{176.0928} & 0.3250 \\
		GANMcC    & 1.8641 & 4672.0123 & 0.2792 \\
		GAN-FM    & 2.1253 & 1312.3091 & \textbf{0.0892} \\
		TIPFNet   & 4.3608 & 2459.1799 & 0.4297 \\
		CMTFusion & 0.6240 & \underline{423.5443} & \underline{0.1777} \\
		MSPFusion & \underline{0.1221} & 574.7442 & 0.6351 \\
		SFMFusion & 1.6083 & 3461.3707 & 3.8373 \\
		Ours      & 1.6479 & 2670.1891 & 3.7940 \\
		\bottomrule
	\end{tabular*}
\end{table}

PolMamba requires 2670.1891~GFLOPs, fewer than RFN-Nest, GANMcC, and SFMFusion, but more than the other five methods. GAN-FM has the shortest average inference time of 0.0892~s, followed by CMTFusion. PolMamba takes 3.7940~s, slightly less than SFMFusion at 3.8373~s but more than the remaining methods. Relative to the compared Mamba-based SFMFusion, PolMamba reduces both FLOPs and runtime. Its overall runtime nevertheless identifies computational efficiency as an important direction for optimization alongside fusion quality.

\subsection{Limitations and Future Work}
The evaluation of PolMamba on AIR-LWIR-Pol and LDDRS examines its use for long-wave infrared polarimetric fusion and downstream vehicle detection. Further work is needed to broaden data coverage, characterize the applicability of the physical prior, extend task validation, and improve computational efficiency.

The physical interpretation of the prior depends on preprocessing and imaging errors. Water-surface images provide a reference field for relative spatial bias compensation across the four channels, but this does not replace full radiometric and polarimetric-response calibration. Grayscale clipping and linear stretching also mean that the input product is no longer strictly a linearly polarized intensity with defined radiometric units. Future work will combine calibration data, channel-response consistency analysis, and registration-error perturbations to establish the prior's operating conditions and develop guidance that accounts for measurement uncertainty.

Downstream validation is currently confined to oriented vehicle detection with the present detector configuration. PolMamba improves precision, $\mathrm{mAP}_{50}$, and $\mathrm{mAP}_{50\text{--}95}$ over both single-modality inputs, but its recall of 0.521 is below the 0.548 obtained with $S_0$. Improvements in average precision therefore do not coincide with increased recall. The aggregate results do not identify the target scales, contrast levels, or backgrounds responsible for missed detections. Future analyses will examine false positives, missed targets, and localization errors by target size, density, and background type, and evaluate other detectors and perception tasks.

The average inference time of 3.7940~s is longer than that of most comparison methods and does not support real-time processing under the current settings. A modest parameter count does not necessarily imply low runtime cost. Runtime profiling will identify the main latency and memory costs of multidirectional scanning, feature rearrangement, and cross-modal interaction. We will investigate reducing redundant scans, optimizing data access, and simplifying the architecture, and assess the effects on fusion and detection under identical hardware, input dimensions, and numerical precision.

\section{Conclusion}\label{sec:conclusion}
PolMamba addresses polarization noise and incomplete intensity--polarization integration through physics-guided prior enhancement and multidirectional Cross-Mamba fusion. A dual-branch LLRR encoder extracts modality features; PPREM combines polarization-response support with spatial structural continuity; and its guidance regulates cross-modal feature propagation. Joint optimization uses prior-weighted multi-window structural similarity, encoder--decoder reconstruction, and a relative lower-bound constraint for polarization saliency. On AIR-LWIR-Pol, PolMamba ranks first in EN, SD, SSIM, and NIMA and second in QCB and CEIQ. Qualitative results show preservation of background textures and local polarization information. Downstream vehicle detection achieves precision of 0.607, $\mathrm{mAP}_{50}$ of 0.574, $\mathrm{mAP}_{50\text{--}95}$ of 0.303, and F1 of 0.561, supporting the use of the fused images for remote sensing vehicle detection against complex backgrounds. Future work will examine cross-scene generalization, interpretable polarization prior estimation, and more direct joint optimization of fusion and downstream perception.

\section*{Acknowledgments}
This research was funded by the Science and Disruptive Technology Program, AIRCAS (No. 2024-AIRCAS-SDTP-11).


\end{document}